\documentclass[runningheads]{llncs}
\usepackage{graphicx}
\usepackage{comment}
\usepackage{amsmath,amssymb} 
\usepackage{color}
\usepackage{enumitem}
\usepackage{hyperref}
\usepackage{times}
\usepackage{epsfig}
\usepackage[ruled,linesnumbered]{algorithm2e}
\usepackage{setspace}
\usepackage{mathtools}
\usepackage{multirow}
\usepackage{caption}
\usepackage{makecell}
\usepackage{tabularx,booktabs}

\usepackage{lipsum}
\usepackage{bm,array}
\usepackage{breakcites}

\def\etal{et~al.}

\makeatletter
\newcommand{\appendixhead}%
{\begin{center}
  {\huge Appendix \small }  
\end{center}}%
\makeatother
\hypersetup{
    colorlinks=true,
    urlcolor=magenta,
}

\begin{document}
\pagestyle{headings}
\mainmatter
\def\ECCVSubNumber{5849}  

\title{Attributional Robustness Training using Input-Gradient Spatial Alignment} 

\titlerunning{ECCV-20 submission ID \ECCVSubNumber} 
\authorrunning{ECCV-20 submission ID \ECCVSubNumber} 
\author{Anonymous ECCV submission}
\institute{Paper ID \ECCVSubNumber}

\titlerunning{Attributional Robustness Training using Input-Gradient Spatial Alignment}
%
\author{Mayank Singh\inst{1}\thanks{Equal contribution} \and
Nupur Kumari\inst{1}$^{*}$ \and
Puneet Mangla\inst{2}\and
Abhishek Sinha\inst{1}\thanks{Work done at Adobe}\and \\
Vineeth N Balasubramanian\inst{2} \and
Balaji Krishnamurthy\inst{1}
}
\authorrunning{M. Singh et al.}
%
\institute{Media and Data Science Research Lab, Adobe, India \\
\email{\{msingh,nupkumar\}@adobe.com, abhishek.sinha94@gmail.com , kbalaji@adobe.com} \and
IIT Hyderabad, India \\
\email{\{cs17btech11029,vineethnb\}@iith.ac.in}
}
\maketitle

\begin{abstract}
Interpretability is an emerging area of research in trustworthy machine learning. Safe deployment of machine learning system mandates that the prediction and its explanation be reliable and robust. Recently, it has been shown that the explanations could be manipulated easily by adding visually imperceptible perturbations to the input while keeping the model's prediction intact. In this work, we study the problem of attributional robustness (i.e. models having robust explanations) by showing an upper bound for attributional vulnerability in terms of spatial correlation between the input image and its explanation map. We propose a training methodology that learns robust features by minimizing this upper bound using soft-margin triplet loss. Our methodology of robust attribution training (\textit{ART}) achieves the new state-of-the-art attributional robustness measure by a margin of $\approx$ 6-18 $\%$ on several standard datasets, ie. SVHN, CIFAR-10 and GTSRB. We further show the utility of the proposed robust training technique (\textit{ART}) in the downstream task of weakly supervised object localization by achieving the new state-of-the-art performance on CUB-200 dataset. Code is available at \href{https://github.com/nupurkmr9/Attributional-Robustness}{https://github.com/nupurkmr9/Attributional-Robustness}.
\end{abstract}

\section{Introduction}
Attribution methods \cite{attr2017grad++,attr2016gradcam,attr2018smoothgrad,attr2013gradient,attr2016inputgradient,attr2017integrated,attr2017deeplift} are an increasingly popular class of explanation techniques that aim to highlight relevant input features responsible for model's prediction. These techniques are extensively used with deep learning models in risk-sensitive and safety-critical applications such as healthcare  \cite{lung,anaemia,eye,skin}, where they provide a human user with visual validation of the features used by the model for predictions. E.g., in computer-assisted diagnosis, \cite{eye} showed that predictions with attribution maps increased accuracy of retina specialists above that of unassisted reader or model alone. Also, in \cite{skin}, the authors improve the analysis of skin lesions by leveraging explanation maps of prediction. 

\begin{figure}[t]
\centering
\scalebox{1.0}{
\includegraphics[width=\textwidth]{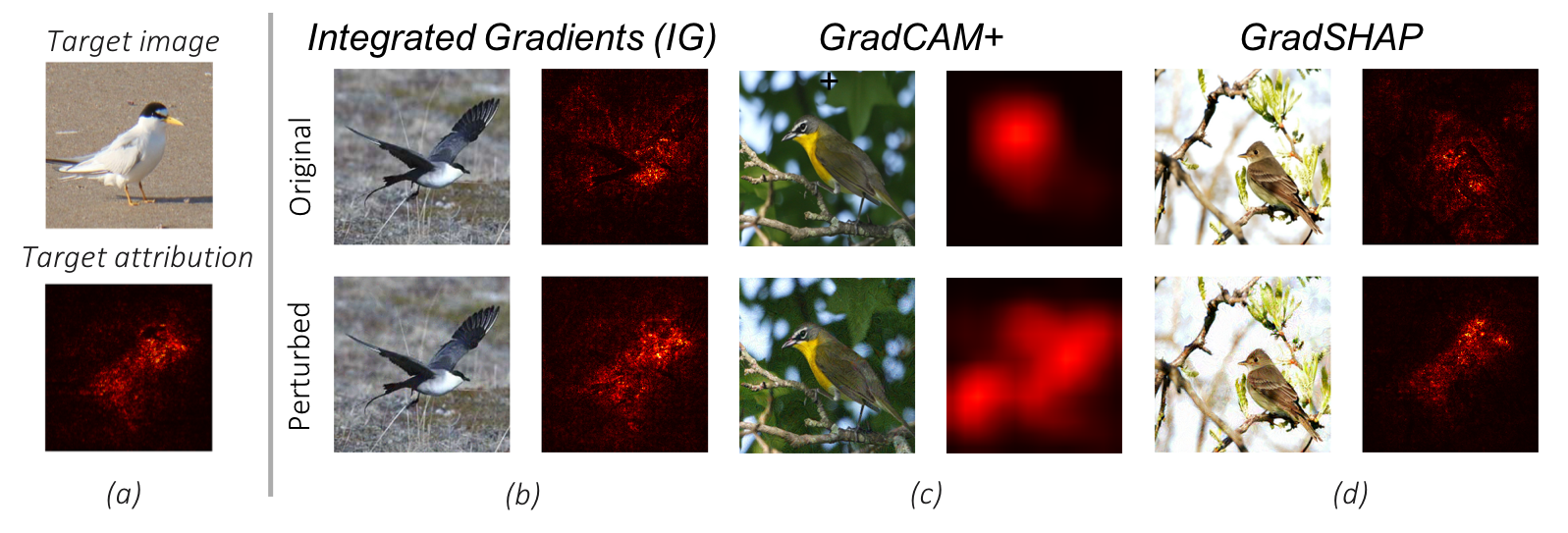}
}
\caption{\footnotesize{Illustration of targeted manipulation \cite{nips_sal} of different attribution maps using the target attribution of (a). Here, (b) Integrated Gradients \cite{attr2017integrated}, (c) GradCAM++ \cite{attr2017grad++} and (d) GradSHAP \cite{gradshap} blocks show : Top (b), (c), (d)  original image and its attribution map; Bottom (b), (c), (d) perturbed image and its attribution map. Both original and perturbed images of (b), (c) and (d) are classified correctly by the ResNet-$50$ trained model on CUB-200 \cite{cub} in the class of Long Tailed Jaeger, Yellow Breasted Chat and Acadian Flycatcher respectively.}}
\label{fig:example_attr}
\end{figure}

It has been recently demonstrated that one could construct targeted \cite{nips_sal} and un-targeted perturbations \cite{aaai_sal,robust_attr_nips_sal} that can arbitrarily manipulate attribution maps without affecting the model's prediction. This issue further weakens the cause of safe application of machine learning algorithms. We show an illustrative example of attribution-based attacks for image classifiers over different attribution methods in Fig. \ref{fig:example_attr}. This vulnerability leads to newer challenges for attribution methods, as well as robust training techniques. The intuition of attributional robustness is that if the inputs are visually indistinguishable with the same model prediction, then interpretation maps should also remain the same. 

As one of the first efforts, \cite{robust_attr_nips_sal} recently proposed a training methodology that aims to obtain models having robust integrated gradient \cite{attr2017integrated} attributions. In addition to being an early effort, the instability
of this training methodology, as discussed in \cite{robust_attr_nips_sal}, limits its usability in the broader context of robust training in computer vision. In this paper, we build upon this work by obtaining an upper bound for attributional vulnerability as a function of spatial correlation between the input image and its explanation map. Furthermore, we also introduce a training technique that minimizes this upper bound to provide attributional robustness. In particular, we introduce a training methodology for attributional robustness that uses soft-margin triplet loss to increase the spatial correlation of input with its attribution map. The triplet loss considers input image as the anchor, gradient of the correct class logit with respect to input as the positive and gradient of the incorrect class with highest logit value with respect to input as the negative. We show empirically how this choice results in learning of robust and interpretable features that help in other downstream weakly supervised tasks.

Existing related efforts in deep learning research are largely focused on robustness to adversarial perturbations \cite{ian2015harness,szegedy2013intriguing}, which are imperceptible perturbations which, when added to input, drastically change the neural network’s prediction. 
While adversarial robustness has been explored significantly in recent years, there has been limited progress made on the front of attributional robustness, which we seek to highlight in this work. Our main contributions can be summarized as:
\begin{itemize}
    \item We tackle the problem of attribution vulnerability and provide an upper bound for it as a function of spatial correlation between the input and its attribution map \cite{attr2013gradient}. 
    We then propose \textit{ART}, a new training method that aims to minimize this bound to learn attributionally robust model. 
    \item Our method outperforms prior work in this direction, and achieves state-of-the-art attributional robustness on Integrated Gradient \cite{attr2017integrated} based attribution method.
    
    \item We empirically show that the proposed methodology also induces immunity to adversarial perturbations and common perturbations \cite{hendrycks2019robustness} on standard vision datasets that is comparable to the state-of-the-art adversarial training technique \cite{madrypgd}. 

    \item We show the utility of \textit{ART} for other computer vision tasks such as weakly supervised object localization and segmentation. Specifically, \textit{ART} achieves state-of-the-art performance in weakly supervised object localization on CUB-200 \cite{cub} dataset.  
\end{itemize}

\section{Related Work}\label{sec_related_work}
Our work is associated with various recent development made in the field of explanation methods, robustness to input distribution shifts and weakly supervised object localization. We hence describe earlier efforts in each of these directions below.

\paragraph{\textbf{Visual Explanation Methods:}}
Various explanation methods have been proposed that focus on producing posterior explanations for the model's decisions. A popular approach to do so is to attribute the predictions to the set of input features \cite{attr2013gradient,attr2015deconv,attr2016inputgradient,attr2017integrated,attr2017deeplift,attr2018LRP}. \textit{Sample-based} explanation methods \cite{sample_explain_1,sample_explain_2} leverage previously seen examples to describe the prediction of the model. \textit{Concept-based} explanation techniques \cite{feature_explain_1,feature_explain_2} aim to explain the decision of the model by high-level concepts. There has also been work that explores interpretability as a built-in property of architecture inspired by the characteristics of linear models \cite{self_explain_nips_2018}. \cite{survey_interpretation_1,survey_interpretation_2} provide a survey of interpretation techniques. Another class of explanation methods, commonly referred to as attribution techniques, can be broadly divided into three categories - gradient/back-propagation, propagation and perturbation based methods. Gradient-based methods attribute an importance score for each pixel by using the derivative of a class score with respect to input features \cite{attr2013gradient,attr2016inputgradient,attr2017integrated}. Propagation-based techniques \cite{attr2018LRP,attr2017deeplift,attr2018ebackprop} leverage layer-wise propagation of feature importance to calculate the attribution maps. Perturbation-based interpretation methods generate attribution maps by examining the change in prediction of the model when the input image is perturbed \cite{attr2016perturb_1,attr2018perturb_2,attr2016perturb_3_lime}. In this work, we primarily report results on the attribution method of Integrated Gradients \textit{IG} \cite{attr2017integrated} that satisfies desirable axiomatic properties and was also used in the previous work \cite{robust_attr_nips_sal}.

\paragraph{\textbf{Robustness of Attribution Maps:}}
Recently, there have been a few efforts \cite{sal_fire,aaai_sal,nips_sal,robust_attr_nips_sal,alvarez2018robustness} that have explored the robustness of attribution maps, which we call attributional robustness in this work. The authors of \cite{aaai_sal,nips_sal,sal_fire} study the robustness of a network's attribution maps and show that the attribution maps can be significantly manipulated via imperceptible input perturbations while preserving the classifier's prediction. Recently, Chen, J. \etal \cite{robust_attr_nips_sal} proposed a robust attribution training methodology, which is one of the first attempts at making an image classification model attributionally robust and is the current state of the art. The method minimizes the norm of difference in Integrated Gradients \cite{attr2017integrated} of an original and perturbed image during training to achieve attributional robustness. In this work, we approach the problem from a different perspective of maintaining spatial alignment between an image and its saliency map. 

\paragraph{\textbf{Adversarial Perturbation and Robustness:}}
Adversarial attacks can be broadly categorized into two types: White-box \cite{seyed2016deepfool,madrypgd,carlini2017towards,sal_adv_structure} and Black-box attacks \cite{blackbox4,spsa_blackbox,alexey2017blackbox,papernot2017blackbox1}. Several proposed defense techniques have been shown to be ineffective to adaptive adversarial attacks \cite{obfuscate,break_alp,carlini2017towards,carlini2019evaluating}. Adversarial training \cite{goodfellow2014explaining,madrypgd,sinha2019harnessing}, which is a defense technique that continuously augments the data with adversarial examples while training, is largely considered the current state-of-the-art to achieve adversarial robustness. \cite{Zhang2019trades} characterizes the trade-off between accuracy and robustness for classification problems and propose a regularized adversarial training method. Recent work of \cite{curvature_2} proposes a regularizer that encourages the loss to behave linearly in the vicinity of the training data, and \cite{feature_fb_scatter} improves the adversarial training by also minimizing the convolutional feature distance between the perturbed and clean examples. Prior works have also attempted to improve adversarial robustness using gradient regularization that minimizes the Frobenius norm of the Hessian of the classification loss with respect to input\cite{hessian2018aaai,hessian2019cure,hessian2015unified} or weights \cite{hessian2018jacobian}. For a comprehensive review of the work done in the area of adversarial examples, please refer \cite{reviewpaper,reviewpaper2}. We show in our work that in addition to providing attributional robustness, our proposed method helps in achieving significant performance improvement on downstream tasks such as weakly supervised object localization. We hence briefly discuss earlier efforts on this task below.

\paragraph{\textbf{Weakly Supervised Object Localization (WSOL):}}
The problem of WSOL aims to identify the location of the object in a scene using only image-level labels, and without any location annotations. Generally, rich labeled data is scarcely available, and its collection is expensive and time-consuming. Learning from weak supervision is hence promising as it requires less rich labels and has the potential to scale. A common problem with most previous approaches is that the model only identifies the most discriminative part of the object rather than the complete object. For example, in the case of a bird, the model may rely on the beak region for classification than the entire bird's shape. In WSOL task, ADL \cite{adl}, the current state-of-the-art method, uses an attention-based dropout layer while training the model that promotes the classification model to also focus on less discriminative parts of the image. For getting the bounding box from the model, ADL and similar other techniques in this domain first extract attribution maps, generally CAM-based\cite{attr2016cam}, for each image and then fit a bounding box as described in \cite{attr2016cam}. We now present our methodology.

\section{Attributional Robustness Training: Methodology}
\label{sec_method}
Given an input image $x \in [0, 1]^n$ with true label $y\in \{1...k\}$, we consider a neural network model $f_{\theta}: \mathbb{R}^n \rightarrow \mathbb{R}^k$ with ReLU activation function that classifies $x$ into one of $k$ classes as $\arg \max f(x)_i$ where $i \in \{1...k\}$. Here, $f(x)_i$ is the $i^{th}$ logit of $f(x)$. Attribution map $A(x,f(x)_i):\mathbb{R}^n \rightarrow \mathbb{R}^n$ with respect to a given class $i$ assigns an importance score to each input pixel of $x$ based on its relevance to the model for predicting the class $i$. 

\subsection{Attribution Manipulation}
It was shown recently \cite{nips_sal,aaai_sal} that for standard models $f_{\theta}$, it is possible to manipulate the attribution map $A(x,f(x)_y)$ (denoted as $A(x)$ for simplicity in the rest of the paper) with visually imperceptible perturbation $\delta$ in the input by optimizing the following loss function.  
\begin{equation}
    \begin{aligned}
    \underset{\delta \in B_{\epsilon}}{\arg \max} \; & D[A(x+\delta,f(x+\delta)_y), A(x,f(x)_y)] \\
    \textnormal{subject to: } & {\arg\max}(f(x)) = {\arg\max}(f(x+\delta))=y
    \end{aligned}
    \label{attr_def}
\end{equation}
where $B_{\epsilon}$ is an $l_p$ ball of radius $\epsilon$ centered at $x$ and $D$ is a dissimilarity function to measure the change between attribution maps. The manipulation was shown for various perturbation-based and gradient-based attribution methods. 

This vulnerability in neural network-based classification models suggests that the model relies on features different from what humans perceive as important for its prediction. The goal of attributional robustness is to mitigate this vulnerability and ensure that attribution maps of two visually indistinguishable images are also nearly identical. In the next section, we propose a new training methodology for attributional robustness motivated from the observation that feature importance in image space has a high spatial correlation with the input image for robust models \cite{odds_madry,icml_theory_sal}.

\subsection{Attributional Robustness Training (\textit{ART})}\label{art}
Given an input image $x \in \mathbb{R}^{n}$ with ground truth label $y \in \{1...k\}$ and a classification model $f_{\theta}$, the gradient-based feature importance score is defined as $\nabla_x f(x)_i: i \in \{1...k\}$ and denoted as $g^{i}(x)$ in the rest of the paper. For achieving attributional robustness, we need to minimize the attribution vulnerability to attacks as defined in Equation \ref{attr_def}. Attribution vulnerability can be formulated as the maximum possible change in $g^y(x)$ in a $\epsilon$-neighborhood of $x$ if $A$ is taken as gradient attribution method \cite{attr2013gradient} and $D$ is a distance measure in some norm $||.||$ i.e.
\begin{equation}
\begin{aligned}
   &  \underset{\delta \in B_{\epsilon}}{\max} || g^y(x+\delta) - g^y(x) || 
\end{aligned}
\label{attr_vul}
\end{equation}

\noindent We show that Equation \ref{attr_vul} is upper bounded by the maximum of the distance between $g^y(x+\delta)$ and $x+\delta$ for $\delta$ in $\epsilon$ neighbourhood of $x$.
\begin{equation}
\footnotesize
\begin{aligned}
     || g^{y}(x +\delta) - g^y(x) || 
     = \;\; & || g^{y}(x +\delta) - (x + \delta) - (g^y(x) -x)  + \delta|| \\
      \leq \;\; &  || g^{y}(x +\delta) - (x + \delta) || +  || g^y(x) -x || + ||\delta|| \\
      \leq \;\; &  || g^{y}(x +\delta) - (x + \delta) || +  \underset{\delta \in B_{\epsilon}}{\max} || g^{y}(x+\delta) - (x+\delta) || + ||\delta||
\end{aligned}
\end{equation}
Taking max on both sides:
\begin{equation}
\begin{aligned}
     \underset{\delta \in B_{\epsilon}}{\max} || g^y(x+\delta)-g^y(x)) ||
    \leq \;\; & 2 \; \underset{\delta \in B_{\epsilon}}{\max} || g^{y}(x + \delta) - (x + \delta) || + ||\epsilon||
\end{aligned}\label{art_proof}
\end{equation}

Leveraging existing understanding \cite{triplet1,triplet} that minimizing the distance between two quantities can benefit from a negative anchor, we use a triplet loss formulation as defined in Equation \ref{l_attr} with image $x$ as an anchor, $g^{y}(x)$ as positive sample and $g^{i^{*}}(x)$ as negative sample. More details about the selection of the optimization objective \ref{l_attr} and choice for the negative sample can be found in Appendix \ref{choice_of_loss}. Hence to achieve attributional robustness, we propose a training technique \textit{ART} that encourages high spatial correlation between $g^{y}(x)$ and $x$ by optimizing $L_{attr}$ which is a triplet loss \cite{triplet} with soft margin on cosine distance between $g^{i}(x)$ and $x$ i.e.
\begin{equation}
    \begin{aligned}
    & L_{attr}(x,y)  =  \log \Big( 1 + \exp \big(- (d(g^{i^{*}}(x),x) - d(g^{y}(x),x) ) \big)\Big)  \\
    \textnormal{where } & d(g^{i}(x),x)  = 1 - \frac{g^{i}(x).x}{||g^{i}(x)||_{2}.||x||_2} \;; 
    \;\; i^{*} = \underset{i \neq y}{\arg \max}f(x)_i 
    \end{aligned}
    \label{l_attr}
\end{equation}
Hence, the classification training objective for \textit{ART} methodology is:\begin{equation}
\begin{aligned}
    & \underset{\theta}{\textnormal{minimize}}\underset{(x,y )}{\mathbb{E}} \Big[ L_{ce}(x+\delta, y) + \lambda \; L_{attr}(x+\delta, y)\Big] \\
    & \textnormal{where } \delta = \underset{||\delta||_{\infty} < \epsilon}{\arg \max} \; L_{attr}(x+\delta,y)
\end{aligned}
\end{equation}\label{eq:art}
Here $L_{ce}$ is the standard cross-entropy loss. The optimization of $L_{attr}$ involves computing gradient of $f(x)_i$ with respect to input $x$ which suffers from the problem of vanishing second derivative in case of ReLU activation, i.e. $\partial^2 f_{i}/\partial x^2$ $\approx$ $0$. To alleviate this, following previous works \cite{nips_sal,robust_attr_nips_sal}, we replace ReLU with softplus non-linearities while optimizing $L_{attr}$ as it has a well-defined second derivative. The softplus approximates to ReLU as the value of $\beta$ in $\textit{softplus}_{\beta}(x) = \frac{log(1+ e^{\beta x})}{\beta}$ increases. Note that optimization of $L_{ce}$ follows the usual ReLU activation pathway. Thus, our training methodology consists of two steps: first, we calculate a perturbed image $\Tilde{x} = x + \delta$ that maximizes $L_{attr}$ through iterative projected gradient descent; secondly, we use $\Tilde{x}$ as the training point on which $L_{ce}$ and $L_{attr}$ is minimized with their relative weightage controlled by the hyper-parameter $\lambda$.

\begin{figure}[t]
\centering
\scalebox{0.6}{
\includegraphics[width=\textwidth,height=\textheight,keepaspectratio]{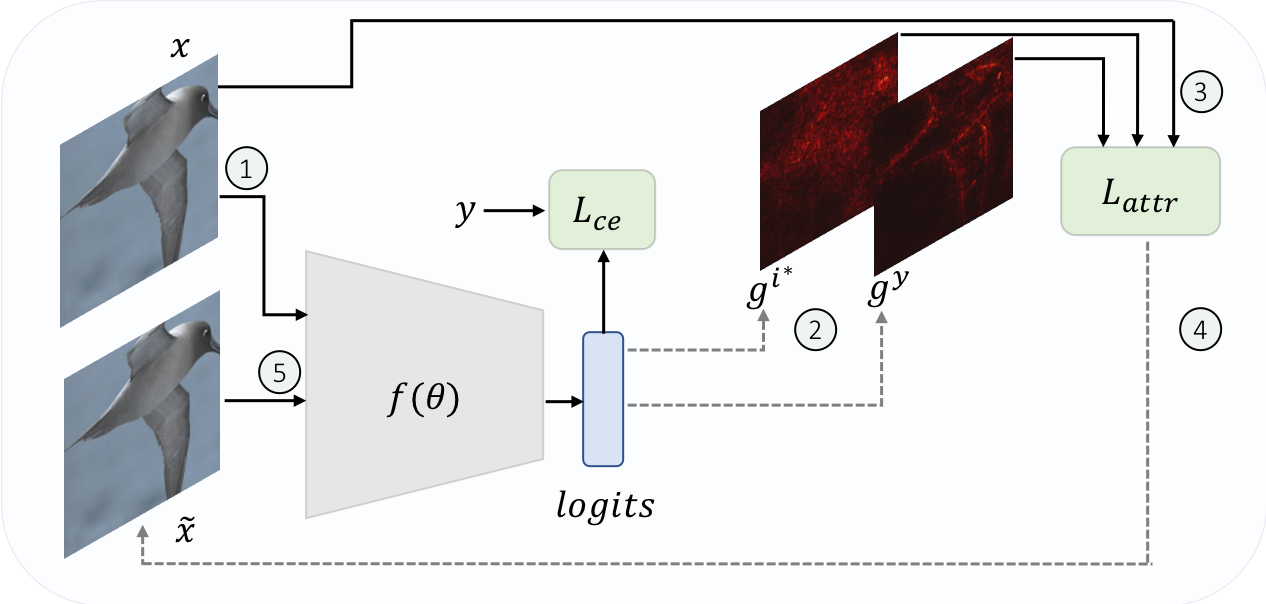}
}
\caption{\footnotesize{Block diagram summarizing our training technique for \textit{ART}. Dashed line represents  backward gradient flow, and bold lines denotes forward pass of the neural network.}}
\label{fig:block-diagram}
\end{figure}

\SetAlgoLined
\begin{algorithm}[!t]
\setstretch{0.9}
\footnotesize
\textbf{Input}: Classification model $f_{\theta}$, training data $X = \{(x_i,y_i)\}$, batch size $b$, number of epochs $E$, number of attack steps $a$, step-size for iterative perturbation $\alpha$, softplus parameter $\beta$, weight of $L_{attr}$ loss $\lambda$. 
    
\For{$epoch \in \{1, 2, ... , E\}$}
    {   Get mini-batch $x,y = \{(x_1,y_1) ... (x_b,y_b)\}$\\
        $\Tilde{x} = x + Uniform[-\epsilon , +\epsilon]$ \\
        \For{i=1,2, ... , $a$}{
            $\Tilde{x} = \Tilde{x} + \alpha*sign(\nabla_{x}L_{attr}({\Tilde{x}},y)) $ \\
            $\Tilde{x} = Proj_{\ell_{\infty}}(\Tilde{x} )$
        }
        

        $i^* = \underset{i \neq y}{\arg \max} \; f(x)_i$ 
        
        Calculate $g^y(\Tilde{x}) = \nabla_{{x}}f(\Tilde{x})_y$ 
        
        Calculate $g^{i^{*}}(\Tilde{x}) = \nabla_{{x}}f(\Tilde{x})_{i^{*}} $\tcp*[r]{\footnotesize{We calculate $g^y(\Tilde{x})$ and $g^{i^{*}}(\Tilde{x})$ using $\textit{softplus}_{\beta}$ activation as described in Section 3.2}} 
    
        $loss = L_{ce}(\Tilde{x},y) +
        \lambda \cdot L_{attr}(\Tilde{x},y)
        $
        
    
        Update $\theta $ using $loss$
    }
\textbf{return} $f_{\theta}$.
\caption{\footnotesize{Attributional Robustness Training (\textit{ART}) }}
\label{train_sal}
\end{algorithm}

Note that the square root of cosine distance for unit $l_2$ norm vectors as used in our formulation of $L_{attr}$ is a valid distance metric and is related to the Euclidean distance as shown in Appendix \ref{cosine}. Through experiments, we empirically show that minimizing the upper bound in Equation \ref{art_proof} as our training objective increases the attributional robustness of the model by a significant margin. The block diagram for our training methodology is shown in Fig \ref{fig:block-diagram}, and its pseudo-code is given in Algorithm \ref{train_sal}.

\subsection{Connection to Adversarial Robustness}\label{connection}
For a given input image $x$, an adversarial example is a slightly perturbed image $x'$ such that $||x-x'||$ is small in some norm but the model $f_{\theta}$ classifies $x'$ incorrectly. Adversarial examples are calculated by optimizing a loss function $L$ which is large when $f(x) \neq y$:
\begin{equation}
    x_{adv} = \underset{{x':||x'-x||_p < \epsilon }}{\arg\max} L(\theta, x', y) 
\end{equation}\label{xadv}
\noindent where $L$ can be the cross-entropy loss, for example. For an axiomatic attribution function $A$ which satisfies the completeness axiom 
i.e. $\textstyle \sum_{j=1}^{n} A(x)_j  = f(x)_y$, it can be shown that $|f(x)_y - f(x')_y| < || A(x) - A(x') ||_1$, as below:   
\begin{equation}
\begin{aligned}
     |f(x)_y - f(x')_y| =&  |\textstyle \sum_{j=1}^{n} A(x)_j - \sum_{j=1}^{n} A(x')_j| \\ 
    \leq & \textstyle \sum_{j=1}^{n} | A(x)_j - A(x')_j| \\
    = & || A(x) - A(x') ||_1
\end{aligned}
\label{ig_adv_proof}
\end{equation} 

The above relationship connects adversarial robustness to attributional robustness as the maximum change in $f(x)_y$ is upper bounded by the maximum change in attribution map of $x$ in its $\epsilon$ neighborhood. Also, it was shown \cite{odds_madry} recently that for an adversarially robust model, gradient-based feature importance map $g^{y}(x)$ has high spatial correlation with the image $x$ and it highlights the perceptually relevant features of the image. For classifiers with a locally affine approximation like a DNN with ReLU activations, Etmann \etal \cite{icml_theory_sal} establish theoretical connection between adversarial robustness, and the correlation of $g^{y}(x)$ with image $x$. \cite{icml_theory_sal} shows that for a given image $x$, its distance to the nearest distance boundary is upper-bounded by the dot product between $x$ and $g^{y}(x)$. The authors of \cite{icml_theory_sal} showed that increasing adversarial robustness increases the correlation between $g^y(x)$ and $x$. Moreover, this correlation is related to the increase in attributional robustness of model as we show in Section \ref{art}.

\subsection{Downstream Task: Weakly supervised Object localization (WSOL)}
As an additional benefit of our approach, we show its improved performance on a downstream task - Weakly supervised Object localization (WSOL), in this case. The problem of WSOL deals with detecting objects where only class label information of images is available, and the ground truth bounding box location is inaccessible. Generally, the pipeline for obtaining bounding box locations in WSOL relies on attribution maps. Also, the task of object detection is widely used to validate the quality of attribution maps empirically. Since our proposed training methodology \textit{ART} promotes attribution map to be invariant to small perturbations in input, it leads to better attribution maps identifying the complete object instead of focusing on only the most discriminative part of the object. We validate this empirically by using attribution maps obtained from our model for bounding-box detection on the CUB dataset and obtaining new state-of-the-art localization results. 

\begin{table}[t]
\caption{\footnotesize{Attributional and adversarial robustness of different approaches on various datasets. Hyper-parameters for attributional attack are same as \cite{robust_attr_nips_sal}. Similarity measures used are IN:\textit{Top-k intersection}, K:\textit{kendall's tau rank order correlation}. The values denote similarity between attribution maps of original and perturbed examples \cite{aaai_sal} based on \textit{Intergrated Gradient} method. }}

\label{sal-robust}
\centering
\scalebox{0.85}{
\begin{tabular}{|c|c|>{\centering}p{1.6cm} | >{\centering}p{1.6cm} | >{\centering}p{1.6cm} | c|}
\hline
\multirow{2}{*}{\textbf{Dataset}} & \multirow{2}{*}{\textbf{Approach}} &  \multicolumn{2}{c|}{\textbf{Attributional Robustness}} &
\multicolumn{2}{c|}{\textbf{Accuracy}}  \\ 
& & IN  & K  & Natural & PGD-40 Attack   
\\ \hline \hline

\multirow{3}{*}{CIFAR-10} & 
 Natural &   40.25 & 49.17 & 95.26  & 0. 
\\\cline{2-6}
& PGD-10 \cite{madrypgd} &   69.00 & 72.27  & 87.32  & 44.07 
\\\cline{2-6}
& ART  & \textbf{92.90} & \textbf{91.76} & 89.84  & 37.58
\\ \hline \hline

\multirow{3}{*}{SVHN} & 
Natural & 60.43  & 56.50  & 95.66  & 0.   
\\\cline{2-6}
& PGD-7 \cite{madrypgd} & 39.67  & 55.56 & 92.84 & 50.12
\\\cline{2-6}
& ART  & \textbf{61.37}   & \textbf{72.60} & 95.47 &   43.56
\\ \hline \hline 

\multirow{3}{*}{GTSRB }  
& Natural & 68.74 & 76.48  & 99.43  & 19.9 
\\\cline{2-6}
& IG Norm \cite{robust_attr_nips_sal} &  74.81 &  75.55  & 97.02  & 75.24 
\\\cline{2-6}
& IG-SUM Norm \cite{robust_attr_nips_sal}  &  74.04 & 76.84  & 95.68  & 77.12
\\\cline{2-6}
& PGD-7 \cite{madrypgd} & 86.13 & 88.42 & 98.36  & 87.49 
\\\cline{2-6}
& ART  & \textbf{91.96} & \textbf{89.34}  & 98.47  & 84.66

\\ \hline \hline

\multirow{3}{*}{Flower }  & 
Natural & 38.22  & 56.43  & 93.91  & 0.   
\\\cline{2-6}
& IG Norm \cite{robust_attr_nips_sal}  & 64.68 &   75.91  & 85.29  & 24.26
\\\cline{2-6}
& IG-SUM Norm \cite{robust_attr_nips_sal} & 66.33 &   79.74 & 82.35  & 47.06
\\\cline{2-6}
& PGD-7 \cite{madrypgd} & \textbf{80.84} & 84.14 &  92.64 & 69.85
\\\cline{2-6}
& ART  & 79.84 & \textbf{84.87} & 93.21  & 33.08 
\\ \hline

\end{tabular}
}

\end{table}

\section{Experiments and Results}
In this section, we first describe the implementation details of \textit{ART} and evaluation setting for measuring the attributional and adversarial robustness. We then show the performance of \textit{ART} on the downstream task of weakly supervised image localization task. 

\subsection{Attributional and Adversarial Robustness}

\subsubsection{Baselines:} We compare our training methodology with the following approaches:
\begin{itemize}[topsep=0pt]
    \item \textit{Natural}: Standard training with minimization of cross entropy classification loss.
    \item \textit{PGD-$n$}: Adversarially trained model with $n$-step PGD attack as in \cite{madrypgd}, which is typically used by work in this area \cite{robust_attr_nips_sal}. 
    \item{ \textit{IG Norm} and \textit{IG-SUM Norm} \cite{robust_attr_nips_sal}}: Current state-of-the-art robust attribution training technique.
\end{itemize}
\subsubsection{Datasets and Implementation Details:}
To study the efficacy of our methodology, we benchmark on the following standard vision datasets: CIFAR-10 \cite{krizhevsky2010cifar}, SVHN \cite{svhn}, GTSRB \cite{gtsrb} and Flower \cite{flower}. For CIFAR-10, GTSRB and Flower datasets, we use Wideresnet-28-10 \cite{wrn} model architecture for \textit{Natural}, \textit{PGD-10} and \textit{ART}. For SVHN, we use WideResNet-40-2 \cite{wrn} architecture. We use the perturbation $\epsilon=8/255$ in ${\ell}_{\infty}$-norm for \textit{ART} and \textit{PGD-n} as in \cite{madrypgd,robust_attr_nips_sal}. We use $\lambda = 0.5$, $a = 3$ and $\beta = 50$ for all experiments in the paper. For training, we use SGD optimizer with step-wise learning rate schedule. More details about datasets and training hyper-parameters are given in Appendix \ref{dataset_attr}. 

\subsubsection{Evaluation:} For evaluating attributional robustness, we follow \cite{robust_attr_nips_sal} and present our results with Integrated Gradient ($IG$)-based attribution maps. We show attributional robustness accuracy of \textit{ART} on other attribution methods in Section \ref{attr_other}. $IG$ satisfies several theoretical properties desirable for an attribution method, e.g. sensitivity and completeness axioms and is defined as: 
\begin{equation}
\begin{aligned}
IG(x,f(x)_i) = (x - \overline{x}) \odot \int_{t=0}^{1} \nabla_x f(\overline{x} + t(x-\overline{x}))_i  dt
\end{aligned}
\end{equation}
where $\overline{x}$ is a suitable baseline at which the function prediction is neutral. 
For computing perturbed image $\Tilde{x}$ on which $IG(\Tilde{x})$ changes drastically from $IG(x)$, we perform Iterative Feature Importance Attack (IFIA) proposed by Ghorbani \etal \cite{aaai_sal} with ${\ell}_{\infty}$ bound of $\epsilon=8/255$ as used by previous work \cite{robust_attr_nips_sal}.

For assessing similarity between $A(x)$ and perturbed image $A(\Tilde{x})$, we use \textit{Top-k intersection} ($IN$) and \textit{Kendall's tau coefficient} ($K$) similar to \cite{robust_attr_nips_sal}. \textit{Kendall's tau coefficient} is a measure of similarity of ordering when ranked by values, and therefore is a suitable metric for comparing attribution maps. \textit{Top-k} intersection measures the percentage of common indices in top-k values of attribution map of $x$ and $\Tilde{x}$. We report average of $IN$ and $K$ metric over random $1000$ samples of test-set. More details about the attack methodology and evaluation parameters can be found in Appendix \ref{section::attackmethodology}. For evaluating adversarial robustness, we perform $40$ step PGD attack \cite{madrypgd} using cross-entropy loss with ${\ell}_{\infty}$ bound of $\epsilon=8/255$ and report the model accuracy on adversarial examples. Table \ref{sal-robust} compares attributional and adversarial robustness across different datasets and training approaches. Our proposed approach \textit{ART} achieves state-of-the-art attributional robustness on attribution attacks \cite{aaai_sal} when compared with baselines. We also observe that \textit{ART} consistently achieves higher test accuracy than \cite{madrypgd} and has adversarial robustness significantly greater than that of the \textit{Natural} model.

\begin{figure}[t]
\centering
\scalebox{1.0}{
\includegraphics[width=1.0\textwidth]{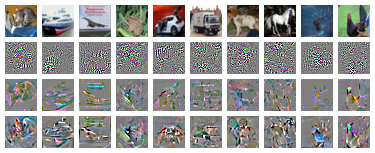}
}
\caption{\footnotesize{Qualitative examples of gradient attribution map \cite{attr2013gradient} for different models on CIFAR-10. Top to bottom: Image; attribution maps for \textit{Natural}, \textit{PGD-10} and \textit{ART} trained models}}
\label{fig:quality}
\end{figure}

\begin{figure}[t]
\centering
\scalebox{1.0}{
\includegraphics[width=1.0\textwidth]{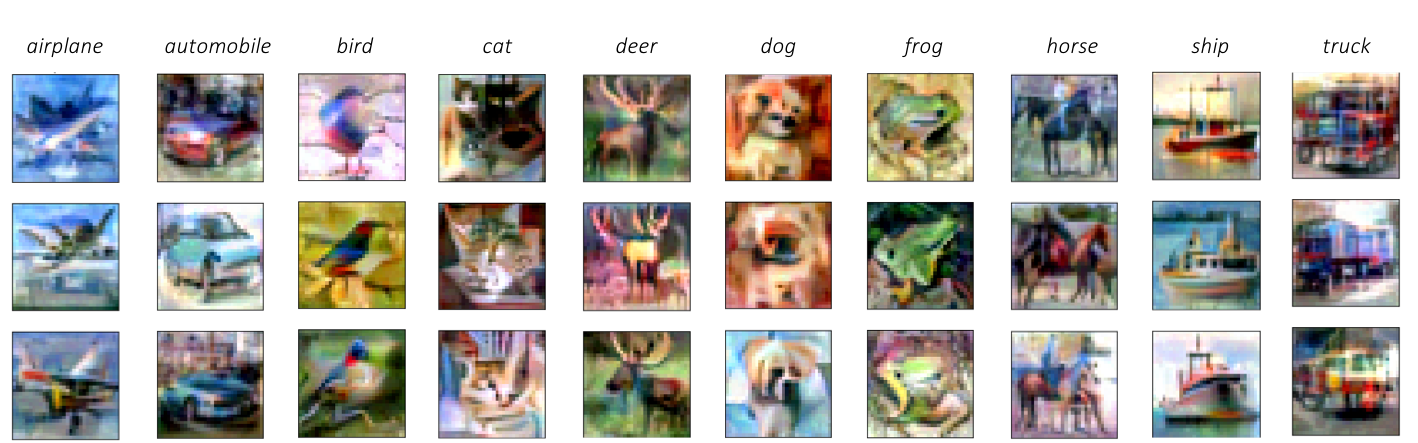}
}
\caption{\footnotesize{Random samples (of resolution $32\times32$) generated using a CIFAR-10 robustly trained \textit{ART} classifier}}
\label{fig:img_gen}
\end{figure}

\subsubsection{Qualitative study of input-gradients for \textit{ART}:}
Motivated by \cite{odds_madry} which claims that adversarially trained models exhibits human-aligned gradients (agree with human saliency), we studied the same with (\textit{ART}), and the results are shown in Fig \ref{fig:quality}. Qualitative study of input-gradients shows a high degree of spatial alignment between the object and the gradient. We also show image generation from random seeds in Fig \ref{fig:img_gen} using robust ART model as done in \cite{santurkar2019synthesis}. The image generation process involves maximization of the class score of the desired class starting from a random seed which is sampled from some class-conditional seed distribution as defined in \cite{santurkar2019synthesis}.

\begin{figure*}[t]
\centering
    \begin{minipage}{.327\textwidth}
      \includegraphics[width=\linewidth]{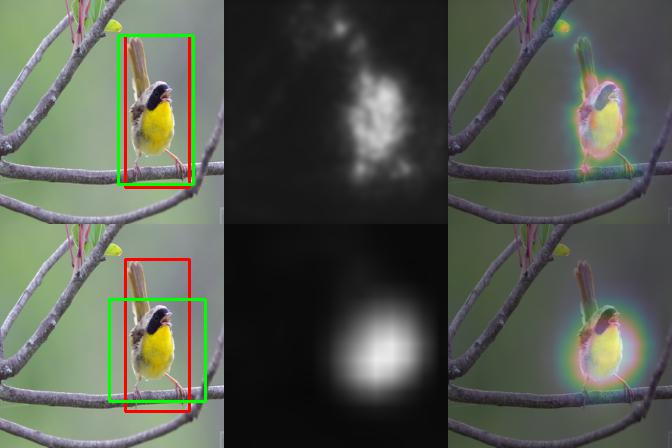}
    \end{minipage}
    \begin{minipage}{.327\textwidth}
      \includegraphics[width=\linewidth]{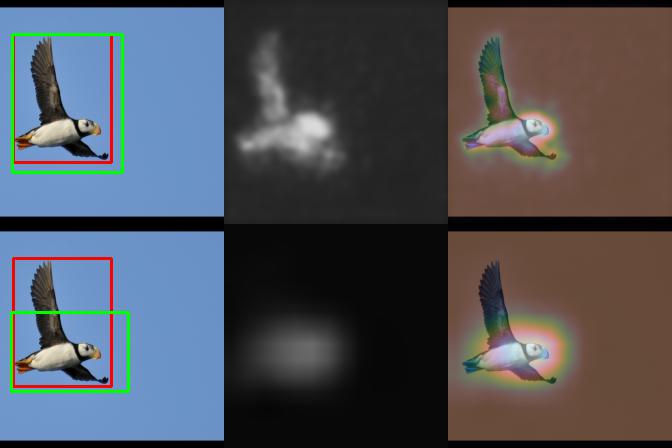}
    \end{minipage}
    \begin{minipage}{.327\textwidth}
      \includegraphics[width=\linewidth]{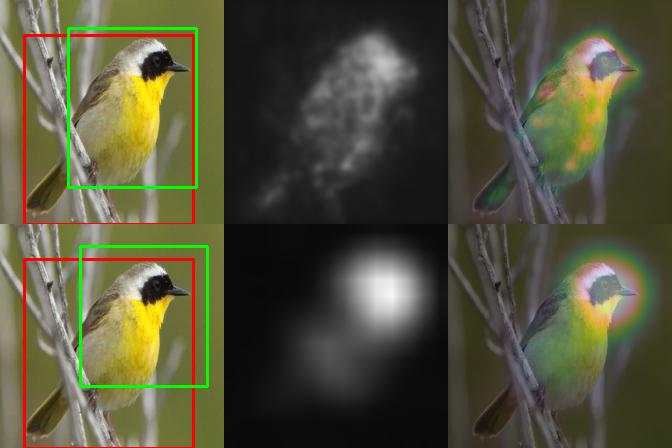}
    \end{minipage}
    \caption{\footnotesize{Comparison of heatmap and estimated bounding box by VGG model trained via our method and ADL on CUB dataset; top row corresponds to our method, and the bottom row corresponds to ADL. The red bounding box is ground truth and green bounding box corresponds to the estimated box}}
    \label{fig:loc}
\end{figure*}

\begin{table}[t]
\caption{\footnotesize{Weakly Supervised Localization on CUB dataset. Bold text refers to the best GT-Known Loc and Top-1 Loc for each model architecture. $*$ denotes directly reported from the paper. $\#$ denotes our implementation of ADL from the official code released by \cite{adl}$^2$ }}
\label{table:weakloc}
\centering
\scalebox{0.8}{
\begin{tabular}{|c|c|c|c|c|c|c|}
\hline
\textbf{Model} & Method & \multicolumn{4}{c|}{\textbf{Saliency Method}} & Top-1 Acc \\
& & \multicolumn{2}{c|}{Grad} & \multicolumn{2}{c|}{CAM} & \\ 
& & GT-Known Loc & Top-1 Loc & GT-Known Loc & Top-1 Loc & \\ \hline 
ResNet50-SE &  ADL \cite{adl} & - & - & - & 62.29$^*$ & 80.34$^*$ \\ \hline
\multirow{4}{*}{ResNet50} &  ADL$^\#$ & 52.93 & 43.78 & 56.85 & 47.53 & 80.0 \\ \cline{2-7}
&  Natural & 50.2 & 42.0 & 60.37 & 50.0 & 81.12 \\ \cline{2-7}
&  PGD-7\cite{madrypgd} & 66.73 & 47.48 & 55.24 & 39.45 & 70.3 \\ \cline{2-7}
&  \textit{ART} & \textbf{82.65} & \textbf{65.22} & 58.87 & 46.02 & 77.51 \\ \hline

\multirow{4}{*}{VGG-GAP} &  ADL$^\#$ & 63.18 & 43.59 & 69.36 & 50.88 & 70.31 \\ \cline{2-7}
&  Natural & 72.54 & 53.81 & 48.75 & 35.03 & 72.94 \\ \cline{2-7}
&  \textit{ART} & \textbf{76.50} & \textbf{57.74} & 52.88 & 40.75 & 74.51 \\ \hline
\end{tabular}}
\end{table}

\subsection{Weakly Supervised Image Localization}\label{sec:wsol}
This task relies on the attribution map obtained from the classification model to estimate a bounding box for objects. We compare our approach with ADL \cite{adl}\footnote{\url{https://github.com/junsukchoe/ADL/tree/master/Pytorch}} on the CUB dataset, which has ground truth bounding box of 5794 bird images. We adopt similar processing steps as ADL for predicting bounding boxes except that we use gradient attribution map $\nabla_{x}f(x)_y$ instead of CAM \cite{attr2016cam}. As a post-processing step, we convert the attribution map to grayscale, normalize it and then apply a mean filtering of $3\times3$ kernel over it. Then a bounding box is fit over this heatmap to localize the object.

We perform experiments on Resnet-50 \cite{resnet} and VGG \cite{vggpaper} architectures. We use ${\ell}_{\infty}$ bound of $\epsilon=2/255$ for \textit{ART} and \textit{PGD-7} training on the CUB dataset. For evaluation, we used similar metrics as in \cite{adl} i.e. \textit{GT-Known Loc}: Intersection over Union (IoU) of estimated box and ground truth bounding box is atleast $0.5$ and ground truth is known; \textit{Top-1 Loc}: prediction is correct and IoU of bounding box is atleast $0.5$; \textit{Top-1 Acc}: top-1 classification accuracy. Details about dataset and training hyper-parameters are given in Appendix \ref{wsol_dataset}. Our approach results in higher \textit{GT-Known Loc} and \textit{Top-1 Loc} for both Resnet-50 and VGG-GAP \cite{adl} model as shown in  Table \ref{table:weakloc}. We also show qualitative comparison of the bounding box estimated by our approach with \cite{adl} in Fig \ref{fig:loc}.

\section{Discussion and Ablation Studies}
\label{sec_ablations}
To understand the scope and impact of the proposed training approach \textit{ART}, we perform various experiments and report these findings in this section. These studies were carried out on the CIFAR-10 dataset.

\noindent\textbf{Robustness to targeted attribution attacks:} In targeted attribution attacks, the aim is to calculate perturbations that minimize dissimilarity between the attribution map of a given image and a target image's attribution map. We evaluate the robustness of \textit{ART} model using targeted attribution attack as proposed in \cite{nips_sal} using the $IG$ attribution method on a batch of $1000$ test examples. To obtain the target attribution maps, we randomly shuffle the examples and then evaluate \textit{ART} and \textit{PGD-10} trained model on these examples. The \textit{kendall's tau coefficient} and \textit{top-k intersection} similarity measure between original and perturbed images on \textit{ART} was $64.76$ and $70.64$ as compared to $36.29$ and $31.81$ on the \textit{PGD-10} adversarially trained model.

\begin{table}[t]
\caption{\footnotesize{Top-1 accuracy of different models on perturbed variants of test-set (GN:Gaussian noise; SN: Shot noise; IN: Impulse noise; DB: Defocus blur; Gl-B: Glass blur; MB: Motion blur; ZB: Zoom blur; S: Snow; F: Fog; B: Brightness; C: Contrast; E: Elastic transform; P: Pixelation noise; J: JPEG compression; Sp-N: Speckle Noise)}}
\label{table:robustness}
\centering
\scalebox{0.8}{
\begin{tabular}{|c|c|c|c|c|c|c|c|c|c|c|c|c|c|c|c|}
\hline
\textbf{Models} & \textbf{GN} & \textbf{SN} & \textbf{IN} & \textbf{DB} & \textbf{Gl-B} & \textbf{MB} & \textbf{ZB} & \textbf{S} & \textbf{F} & \textbf{B} & \textbf{C} & \textbf{E} & \textbf{P} & \textbf{J} & \textbf{Sp-N}\\ \hline
Natural  & 49.16 & 61.42 & 59.22 & 83.55 & 53.84 & 79.16 & 79.18 & 84.53 & \textbf{91.6} & \textbf{94.37} & \textbf{87.63} & 84.44 & 74.12 & 79.76 & 65.04 \\ \hline
PGD-10 & 83.32 & 84.33 & 73.73 & 83.09 & 81.27 & 79.60 & 82.07 & 82.68 & 68.81 & 85.97 & 57.86 & 81.68 & 85.56 & 85.56 & 83.64 \\ \hline
\textit{ART} & \textbf{85.44} & \textbf{86.41} & \textbf{77.07} & \textbf{86.07} &  \textbf{81.70} & \textbf{83.14} & \textbf{85.54} & \textbf{84.99} & {71.04} & {89.42} & {56.69} & \textbf{84.72} & \textbf{87.64} & \textbf{87.89} & \textbf{86.02}  \\
\hline

\end{tabular}}
\end{table}

\begin{table}[t]
\centering
    \begin{minipage}[b]{.42\textwidth}
    \caption{\footnotesize{Attributional Robustness on CIFAR-10 for other attribution methods}}
    \label{table:diff_attr}
    \scalebox{0.8}{
    \begin{tabular}{|c|c|c|c|c|}
    \hline
    \multirow{2}{*}{\textbf{Model}} & \multicolumn{2}{c|}{\textbf{Gradient\cite{attr2013gradient}}} &
    \multicolumn{2}{c|}{\textbf{GradSHAP \cite{gradshap}}} \\
      & IN  & K  & IN  & K \\ \hline \hline
    Natural & 13.72 & 9.5 & 4.5 & 16.52\\
    PGD-10 \cite{madrypgd} & 54.8 & 54.06 & 45.05 & 59.80 \\
    \textit{ART} & \textbf{76.07} & \textbf{70.31} & \textbf{48.31} & \textbf{62.35}\\
    \hline
    \end{tabular}}
      
    \end{minipage}
    \hspace{0.7cm}
    \begin{minipage}[t]{.45\textwidth}
      \includegraphics[width=\linewidth,height=0.40\linewidth]{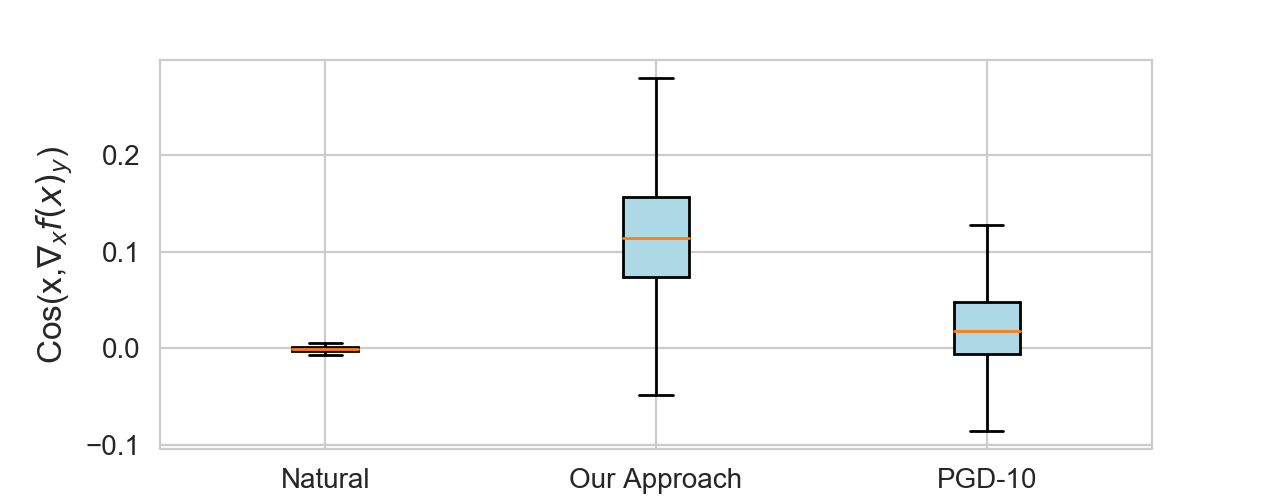}
      \captionof{figure}{\footnotesize{Cosine between $x$ and $\nabla_x f(x)_y$ for different models over test-set of CIFAR-10}}
      \label{fig:var_plot}
    \end{minipage}

\end{table}

\noindent\textbf{Attributional robustness for other attribution methods:} \label{attr_other} We show the efficacy of \textit{ART} against attribution attack \cite{aaai_sal} using gradient\cite{attr2013gradient} and gradSHAP\cite{gradshap} attribution methods in Table \ref{table:diff_attr}. We observe that \textit{ART} achieves higher attributional robustness than \textit{Natural} and \textit{PGD-10} models on \textit{Top-k intersection} (IN) and \textit{Kendall's tau coefficient} (K) measure. We also compare the cosine similarity between $x$ and $g^y(x)$ for all models trained on CIFAR-10 dataset and show its variance plot in Fig. \ref{fig:var_plot}. We can see that \textit{ART} trained model achieves higher cosine similarity \textit{Natural} and \textit{PGD-10} models. This empirically validates that our optimization is effective in increasing the spatial correlation between image and gradient. 

\noindent\textbf{Robustness against gradient-free and stronger attacks:}
To show the absence of gradient masking and obfuscation \cite{obfuscate,carlini2019evaluating}, we evaluate our model on a gradient-free adversarial optimization algorithm \cite{spsa_blackbox} and a stronger PGD attack with a larger number of steps. We observe similar adversarial robustness when we increase the number of steps in PGD-attack. For $100$ step and $500$ step PGD attacks, \textit{ART} achieves $37.42$ $\%$ and $37.18$ $\%$ accuracy respectively. On the gradient-free SPSA \cite{spsa_blackbox} attack, \textit{ART} obtains $44.7$ adversarial accuracy that was evaluated over $1000$ random test samples.

\noindent\textbf{Robustness to common perturbations \cite{hendrycks2019robustness} and spatial adversarial perturbations \cite{spatialrobustness}:} We compare \textit{ART} with \textit{PGD-10}-based adversarially trained model on the common perturbations dataset \cite{hendrycks2019robustness} for CIFAR-10. The dataset consists of perturbed images of $15$ common-place visual perturbations at five levels of severity, resulting in $75$ distinct corruptions. We report the mean accuracy over severity levels for all $15$ types of perturbations and observe that \textit{ART} achieves better generalization than other models on a majority of these perturbations, as shown in Table \ref{table:robustness}. On PGD-40 $\ell_2$ norm attack with $\epsilon=1.0$ and spatial attack \cite{spatialrobustness} we observe robustness of $39.65\%$, $11.13 \%$ for \textit{ART} and $29.68\%$, $6.76\%$ for \textit{PGD-10} trained model, highlighting the improved robustness provided by our method. We show more results on varying $\epsilon$ in adversarial attacks and combining \textit{PGD} adversarial training \cite{madrypgd} with \textit{ART} in Appendix \ref{sec_adv_robustness_ablations}.

\begin{figure}[t]
\centering
\scalebox{.9}{
\includegraphics[width=\textwidth]{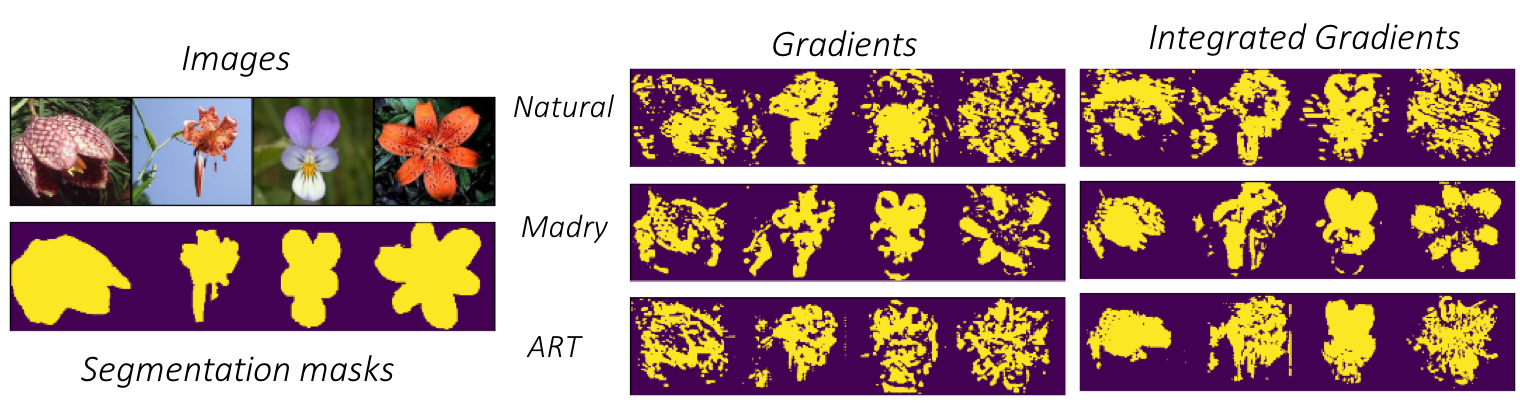}
}
\caption{\footnotesize{Example images of weakly supervised segmentation masks obtained from different models via different attribution methods}}
\label{fig:wsos}
\end{figure}

\noindent\textbf{Image Segmentation:}
Data annotations collection for image segmentation task is time-consuming and costly. Hence, recent efforts \cite{sec,top-down-seg,multitask-seg,superpixel-seg,guided-labeling,graphlet-cut,flower-seg} have focused on weakly supervised segmentation models, where image labels are leveraged instead of segmentation masks. Since models trained via our approach perform well on WSOL, we further evaluate it on weakly supervised image segmentation task for Flowers dataset \cite{flower} where we have access to segmentation masks of 849 images. Samples of weakly-supervised segmentation mask obtained from attribution maps on various models are shown in Fig. \ref{fig:wsos}. We observe that attribution maps of \textit{ART} can serve as a better prior for segmentation masks as compared to other baselines. We evaluate our results using \textit{Top-1 Seg} metric which considers an answer as correct when the model prediction is correct and the IoU betweeen ground-truth mask and estimated mask is atleast $0.5$. We compare \textit{ART} against \textit{Natural} and \textit{PGD-7} trained models using gradient\cite{attr2013gradient} and IG \cite{attr2017integrated} based attribution map. Attribution maps are converted into gray-scale heatmaps and a smoothing filter is applied as a post-processing step. We obtain a \textit{Top-1 Seg} performance of $0.337$, $0.422$, and $0.604$ via IG attribution maps and $0.244$, $0.246$, $0.317$ via gradient maps for \textit{Natural}, \textit{PGD-7} and \textit{ART} models respectively.

\begin{figure}[t]
\centering
\scalebox{0.8}{
\includegraphics[width=0.9\textwidth,height=0.18\textheight]{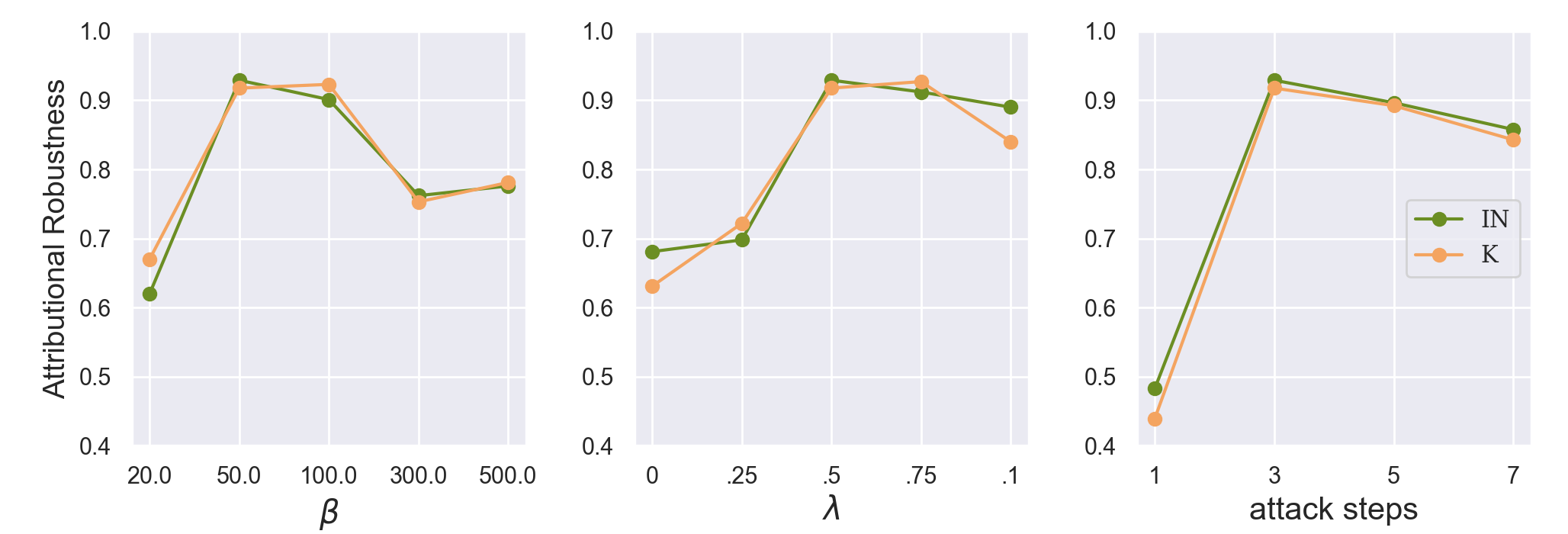}
}
\caption{\footnotesize{Top-k Intersection (IN) and Kendall correlation (K) measure of attributional robustness on varying $\beta$, $\lambda$ and attack steps in our training methodology on CIFAR-10}}
\label{fig:eval_plot}
\end{figure}

\begin{figure}[t]
\centering
\scalebox{0.8}{
\includegraphics[width=0.9\textwidth,height=0.18\textheight]{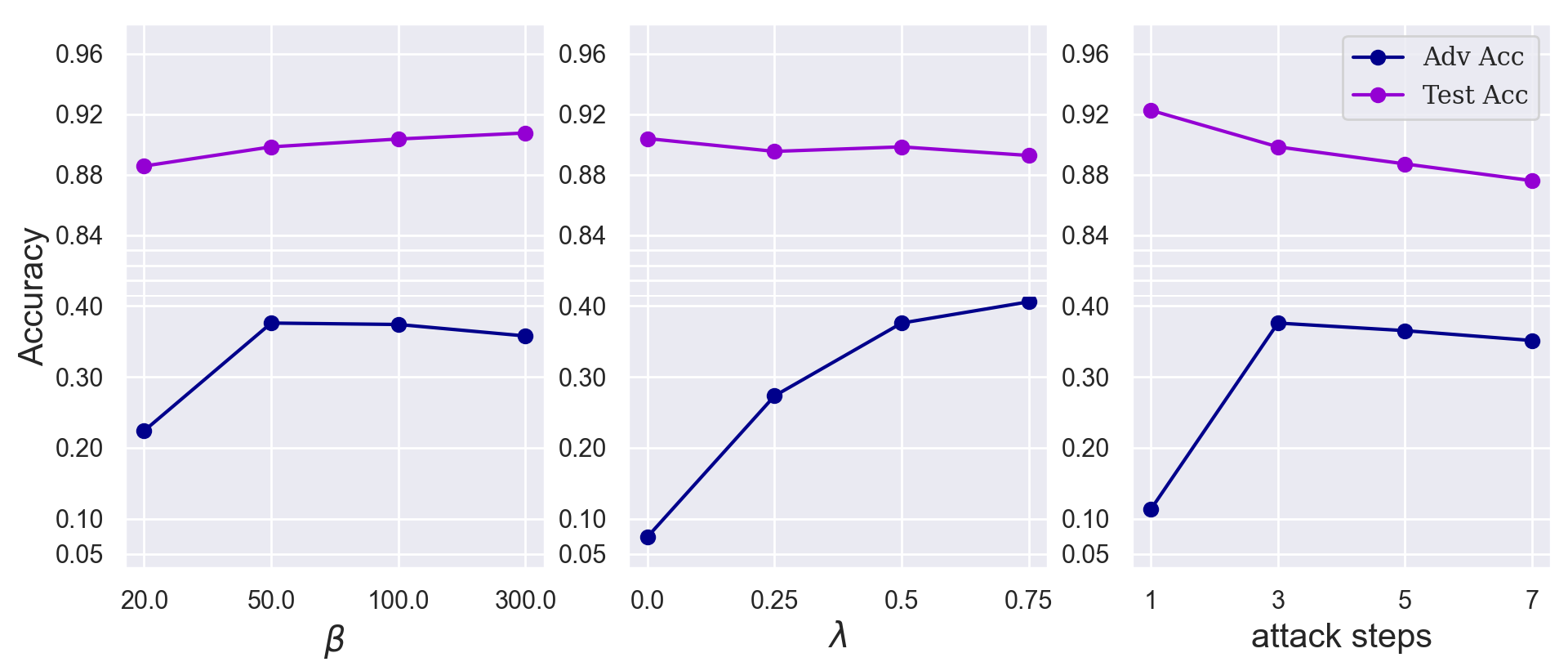}
}
\caption{\footnotesize{Test accuracy and adversarial accuracy (PGD-40 perturbations) on varying $\beta$, $\lambda$ and attack steps in our training methodology on CIFAR-10}}
\label{fig:acc_plot}
\end{figure}

\noindent\textbf{Effect of $\beta$, $\lambda$ and $a$ on performance:}
We perform experiments to study the role of $\beta$, $\lambda$ and $a$ as used in Algorithm \ref{train_sal} on the model performance by varying one parameter and fixing the others on their best-performing values, i.e. $50$, $0.5$ and $3$ respectively. Fig. \ref{fig:eval_plot} shows the plots of attributional robustness. Fig. \ref{fig:acc_plot} shows the plots of test accuracy and adversarial accuracy on ${\ell}_{\infty}$ PGD-40 perturbations with $\epsilon=8/255$ along varying parameters. From Fig. \ref{fig:acc_plot}, we observe that adversarial accuracy initially increases with increasing $\beta$, but the trend reverses for higher values of $\beta$. Similar is the trend for attributional robustness on varying $\beta$ as can be seen from the Fig. \ref{fig:eval_plot}. On varying $\lambda$, we find that the attributional and adversarial robustness of the model increases with increasing $\lambda$ and saturates after $0.75$. However, the test accuracy starts to suffer as the magnitude of $\lambda$ increases. For attack steps parameter $a$, we find that the performance in terms of test accuracy, adversarial accuracy and attributional robustness saturates after $3$ as shown in the right-side plot of Fig. \ref{fig:eval_plot} and Fig. \ref{fig:acc_plot}.

\section{Conclusion}
We propose a new method for the problem space of attributional robustness, using the observation that increasing the alignment between the object in an input and the attribution map generated from the network's prediction leads to improvement in attributional robustness. We empirically showed this for both un-targeted and targeted attribution attacks over several benchmark datasets. We showed that the attributional robustness also brings out other improvements in the network, such as reduced vulnerability to adversarial attacks and common perturbations. For other vision tasks such as weakly supervised object localization, our attributionally robust model achieves a new state-of-the-art accuracy even without being explicitly trained to achieve that objective. We hope that our work can open a broader discussion around notions of robustness and the application of robust features on other downstream tasks.

\noindent\textbf{Acknowledgements.} This work was partly supported by the Ministry of Human Resource Development and Department of Science and Technology, Govt of India through the UAY program.

\clearpage
%
%
\bibliographystyle{splncs04}
\bibliography{5849}

\newpage
\appendixhead

\appendix

\section{Attributional Robustness: Additional Details and Results}
\label{sec_art_additional_results}
In this section, we provide details of the datasets as mentioned in the main paper (Section 4.1), as well as some additional results on attributional robustness.

We qualitatively show in Figure \ref{fig:example-attack} that attribution maps generated via \textit{ART} are robust to attribution manipulation unlike \textit{Natural} model. We also report the Top-$1000$ Intersection and Kendall’s Correlation between original and perturbed saliency maps for \textit{ART} and \textit{Natural} models. We use target attribution attack as mentioned in \cite{nips_sal} to perturb the attributions while keeping the predictions same. For images in Figure \ref{fig:example-attack}, the model predictions are correct and the attribution maps are computed using Integrated Gradient \cite{attr2017integrated}. We observe that attributions of the \textit{Natural} model are visually and quantitatively fragile as attributions are easily manipulated to resemble target attribution map that is present in the rightmost column of the figure. However, it can seen from the figure that \textit{ART} models show high robustness to attribution manipulations. 

\begin{table}[]
\centering
\scalebox{0.98}{
\begin{tabular}{cl}
  \begin{minipage}{.68\textwidth}
      \includegraphics[width=\linewidth,height=0.55\linewidth]{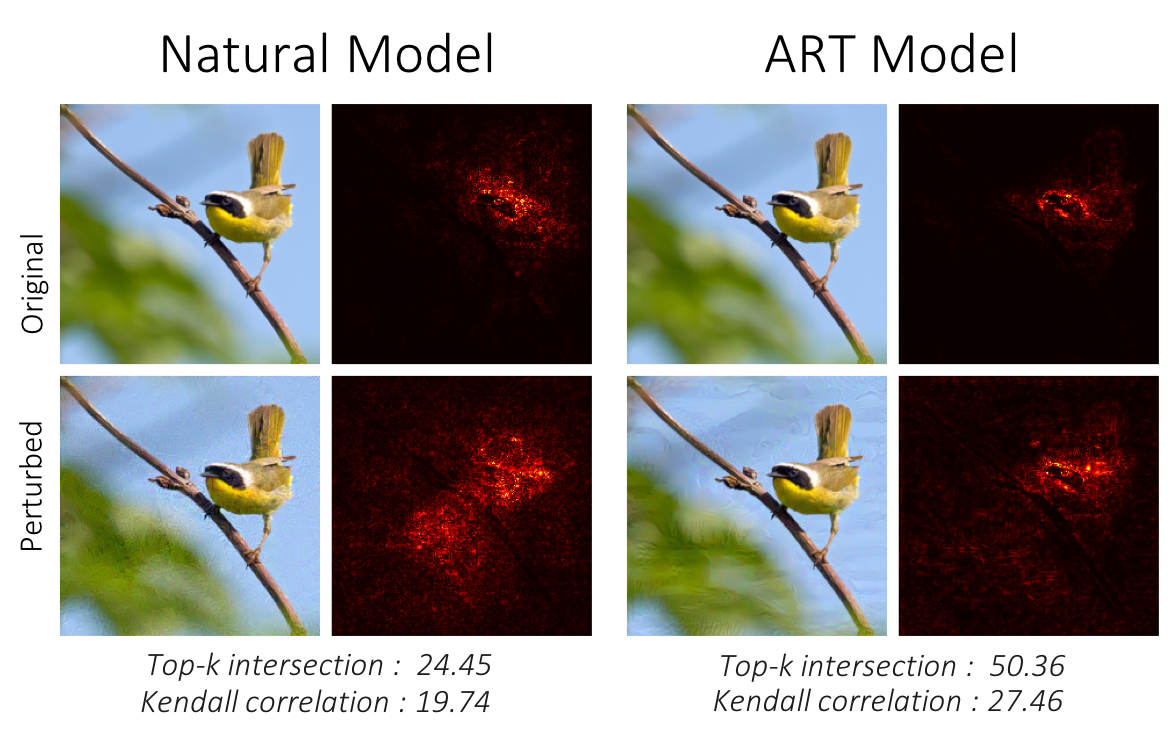}
    \end{minipage} 
  &  \\
  \begin{minipage}{.68\textwidth}
      \includegraphics[width=\linewidth,height=0.55\linewidth]{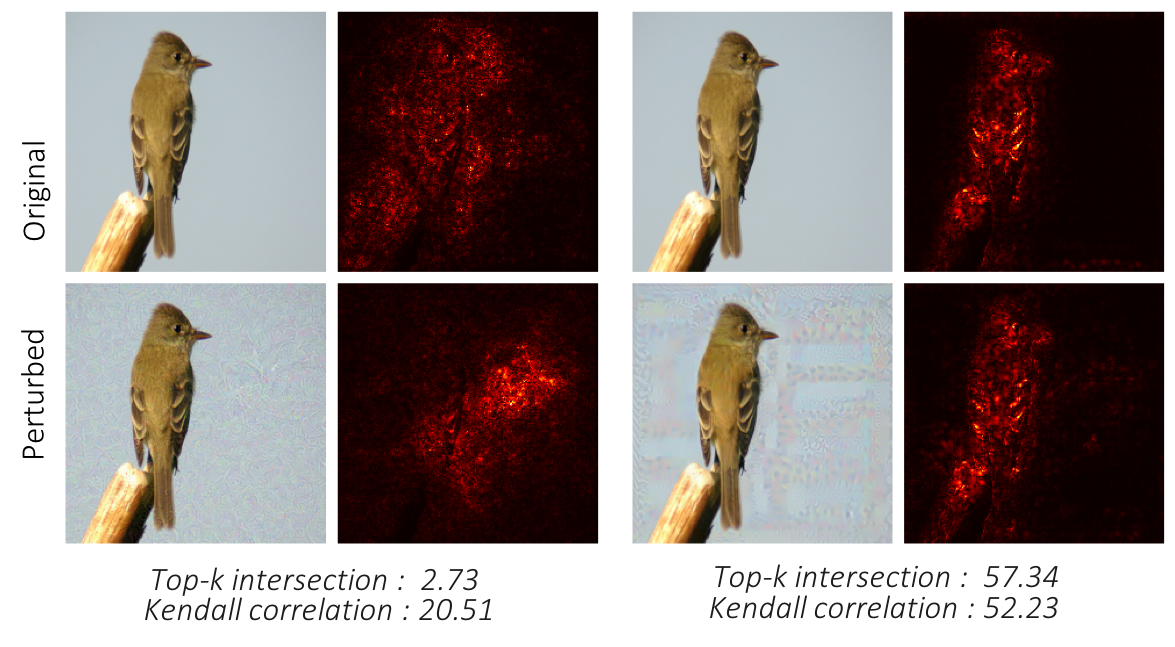}
    \end{minipage}
  & \begin{tabular}[!b]{l} \\ \\ \\ 
    \includegraphics[width=0.17\linewidth,height=0.2\linewidth]{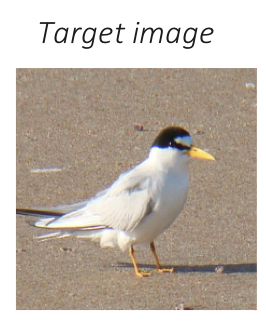}
 \end{tabular} \\ 
 \begin{minipage}{.68\textwidth}
      \includegraphics[width=\linewidth,height=0.55\linewidth]{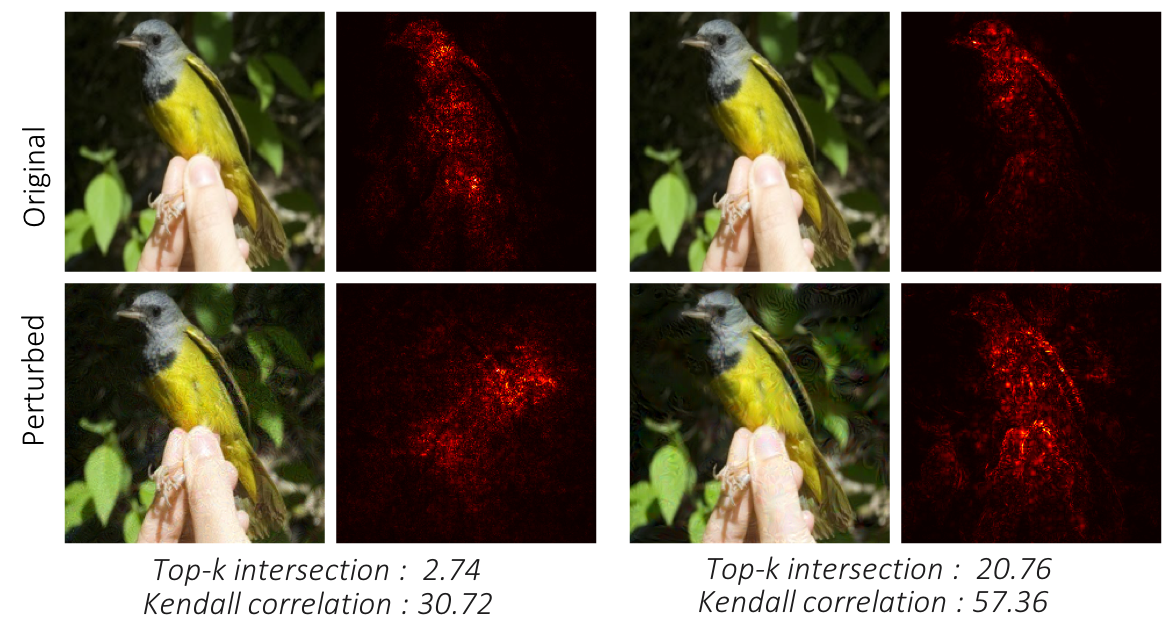}
    \end{minipage}
 & \begin{tabular}[t]{l} 
    \includegraphics[width=0.17\linewidth,height=0.2\linewidth]{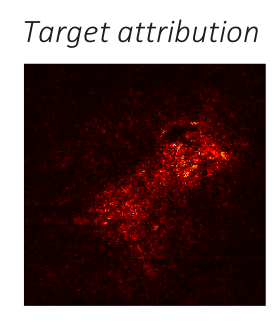}
    \\ 
 \end{tabular} \\
 \begin{minipage}{.68\textwidth}
      \includegraphics[width=\linewidth,height=0.55\linewidth]{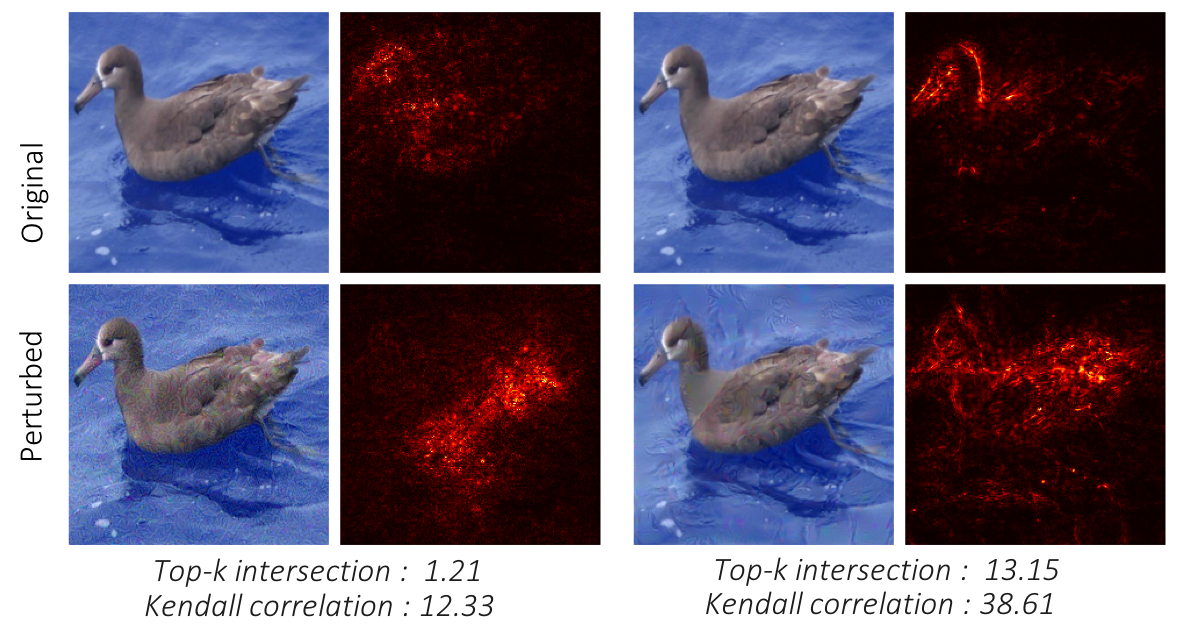}
    \end{minipage} & 
\end{tabular}}
\captionof{figure}{\footnotesize{Targeted attribution attack \cite{nips_sal} using integrated gradient ($IG$) attribution map on \textit{Natural} and \textit{ART} trained model. Top-$1000$ intersection and Kendall correlation between $IG$ attribution map of original and perturbed images is shown below each image. The target attribution manipulation uses the attribution map as depicted in the rightmost column of this figure. }}\label{fig:example-attack}
\end{table}

\subsection{Choice of optimization objective $L_{attr}$ and its variants}\label{choice_of_loss}

Our choice for the loss function was based on the empirical analysis as reported in table \ref{losstype} on CIFAR-10. We empirically observed that instead of directly minimizing $\ell_{2}$ distance between $x$ and $g^y(x)$ in Equation 4 of main paper, cosine distance led to better robustness. We believe this is because cosine avoids scale mismatch issues in $x$ and $g^y(x)$ magnitudes. The triplet loss is only introduced to improve performance on attributional robustness objective. For negative sample selection, we choose $i^*$ as second most likely class, which represents most uncertainty, following standard principles of hard negative mining in triplet loss \cite{triplet,triplet1}. For other choices of $i^*$ , we observed a performance drop.

\begin{table}[]
\centering
\caption{\footnotesize{Comparison of different loss functions used as the objective function for increasing attributional robustness on CIFAR-10 }}
\begin{tabular}{|c|c|c|c|c|}
\hline
\multirow{2}{*}{Optimization Objective}                                                & \multicolumn{2}{c|}{Attributional Robustness} & \multirow{2}{*}{Test Accuracy} & \multirow{2}{*}{\begin{tabular}[c]{@{}c@{}}Adversarial\\ Accuracy\end{tabular}} \\
                                                                                       & \quad \quad IN \qquad \qquad                  & K                     &                                &                                                                                 \\ \hline
Equation 2                                                                             & 74.78                 & 71.40                 & 91.34                          & 15.15                                                                           \\ \hline
Equation 4 : $\ell_{2}$ distance                                                            & 68.41                 & 69.75                 & 91.66                          & 16.64                                                                           \\ \hline
Equation 4 : Cosine distance                                                           & 91.25                 & 89.28                 & 89.21                          & 35.95                                                                           \\ \hline
\begin{tabular}[c]{@{}c@{}}Equation 5 : ART \\ with $i^{*}$=argmin(logit)\end{tabular} & 90.75                 & 83.32                 & 89.94                          & 37.93                                                                           \\ \hline
Equation 5 : ART (ours)                                                                & 92.90                 & 91.76                 & 89.84                          & 37.50                                                                           \\ \hline
\end{tabular}
\label{losstype}
\end{table}

\subsection{Cosine distance in $L_{attr}$ loss}\label{cosine}
Following our discussion in Sec $3.2$ of the main paper, we now elaborate on the relation of cosine distance in a unit $\ell_{2}$-norm surface of vectors with Euclidean distance. We show below that squared Euclidean distance is proportional to the cosine distance for unit $\ell_{2}$ norm space of vectors. Euclidean distance is a valid distance function and follows the triangle inequality which we use in Eqn $3$ for obtaining the upper bound on attributional robustness as a function of the distance between an image and its attribution map.

\noindent Given two vectors $x$ and $\Tilde{x}$, with unit $\ell_{2}$ norm i.e. $||x||_2 = 1$ and $||\Tilde{x}||_2 = 1$, cosine distance between them is related to their Euclidean distance as follows:
\begin{equation}
\begin{aligned}
   (||x-\Tilde{x}||_2)^2 & = (x - \Tilde{x})^\top.(x-\Tilde{x}) \\
   & = x^\top x + \Tilde{x}^\top \Tilde{x} - 2.x^\top.\Tilde{x} \\
   & = ||x||_2 + ||\Tilde{x}||_2 -2.x^\top.\Tilde{x} = 1 + 1 - 2.x^\top.\Tilde{x} \\ 
   & = 2(1 - x^\top.\Tilde{x} ) = 2.\textit{CosineDistance}(x,\Tilde{x})
\end{aligned}
\end{equation}

\subsection{Dataset and Implementation Details}\label{dataset_attr}
Below, we describe the datasets and hyper-parameters used for experiments, which we could not include in the main paper owing to space constraints.

\subsubsection{SVHN}\hspace*{\fill} \\
\noindent \underline{Data and Model:} SVHN dataset \cite{svhn} consists of images of digits obtained from house numbers in Google Street View images, with $73257$ digits for training and $26032$ digits for testing over $10$ classes. We perform experiments on SVHN using WideResNet-40-2 \cite{wrn} architecture for training on reported approaches. \vspace{3pt}

\noindent \underline{Hyperparameters for Training:} \newline
\textit{Natural:} We use SGD optimizer with an initial learning rate of $0.1$, momentum of $0.9$,  $l_2$ weight decay of $2\mathrm{e}{-4}$ and batch size of $256$. We train it for $200$ epochs with a learning rate schedule decay of $0.1$ at $50\textsuperscript{th}$, $80\textsuperscript{th}$ and $0.5$ at $150\textsuperscript{th}$ epoch.
\newline
\textit{PGD-7:} We use the training configuration as in \cite{model_cifar} to perform $7$-step adversarial training with $\epsilon=8/255$ and step size $2.5/255 $.      
\newline 
\textit{ART:} We use the same training configuration as mentioned for \textit{Natural} model, $\beta = 50$ and $\lambda = 0.5$. We calculate $\Tilde{x}$ using $\epsilon=8/255$, step size $1.5/255 $ and number of steps $a = 3$.

\subsubsection{CIFAR-10} \hspace*{\fill} \\
\underline{Data and Model:} CIFAR-10 dataset \cite{krizhevsky2010cifar} consists of $50000$ training images for $10$ classes with resolution of $32\times32\times3$. We normalize the images with its mean and standard deviation for training. We train a WideResNet28-10 \cite{wrn} model for all the experiments on this dataset.
\vspace{3pt}

\noindent \underline{Hyperparameters for Training:} \newline
\textit{Natural:} We use SGD optimizer with an initial learning rate of $0.1$, momentum of $0.9$,  $l_2$ weight decay of $2\mathrm{e}{-4}$ and batch size of $256$. We train it for $100$ epochs with a learning rate schedule decay of $0.1$ at $50\textsuperscript{th}$, $80\textsuperscript{th}$ and $0.5$ at $150\textsuperscript{th}$ epoch.
\newline
\textit{PGD-10:} We use the training configuration as mentioned in \cite{model_cifar} to perform $10$-step adversarial training with $\epsilon=8/255$ and step size $2/255$.
\newline
\textit{ART:} We use the same training configuration as mentioned for \textit{Natural} model with $\beta = 50$ and $\lambda = 0.5$. We calculate $\Tilde{x}$ using $\epsilon=8/255$, step size $1.5/255 $ and number of steps $a = 3$. 

\subsubsection{GTSRB} \hspace*{\fill} \\ 
\underline{Data and Model:} German Traffic Signal Recognition Benchmark \cite{gtsrb} consists of $43$ classes of traffic signals with $34,799$ training images, $4,410$ validation images and $12,630$ test images. We resize the images to $32\times32\times3$ and normalize the images with its mean and standard deviation for training. To balance the number of images for each class, we use data augmentation techniques consisting of rotation, translation, and projection transforms to extend the training set to $10,000$ images per class as in \cite{robust_attr_nips_sal}. We train WideResNet28-10 \cite{wrn} model for carrying out experiments related to this dataset.
\vspace{3pt}

\noindent \underline{Hyperparameters for Training:} \newline
\textit{Natural:} We use SGD optimizer with an initial learning rate of $0.1$, momentum of $0.9$,  $l_2$ weight decay of $2\mathrm{e}{-4}$ and batch size of $128$. We train it for $12$ epochs with a learning rate schedule decay of $0.1$ at $4\textsuperscript{th}$, $6\textsuperscript{th}$ and $0.5$ at $10\textsuperscript{th}$ epoch.
\newline
\textit{PGD-7:} We use the training configuration same as \cite{robust_attr_nips_sal} to perform $7$-step adversarial training with $\epsilon=8/255$ and step size $2/255$.
\newline
\textit{IG Norm} and \textit{IG-Sum Norm \cite{robust_attr_nips_sal}: } We report the accuracy as mentioned in the paper \cite{robust_attr_nips_sal}.
\newline
\textit{ART:} We use the same training configuration as mentioned for \textit{Natural} model with $\beta = 50$ and $\lambda = 0.5$. We calculate $\Tilde{x}$ using $\epsilon=8/255$, step size $1.5/255 $ and number of steps $a = 3$.

\subsubsection{Flower}\hspace*{\fill} \\ 
\underline{Data and Model:} Flower dataset \cite{flower} has $17$ categories with $80$ images for each class. We resize the images to $128\times128\times3$ and normalize it with its mean and standard deviation for training. The training set consists of $1,224$ images with $72$ images per class. The test set compromises of $136$ images with $8$ images per class. We use standard data augmentation techniques of rotation, translation, and projection transforms to extend the training data so that each class contains $1,000$ training examples as proposed in \cite{robust_attr_nips_sal}. We use WideResNet28-10 \cite{wrn} model for the reported approaches.
\vspace{3pt}

\noindent \underline{Hyperparameters for Training:} \newline
\textit{Natural:} We use SGD optimizer with an initial learning rate of $0.1$, momentum of $0.9$, $l_2$ weight decay of $2\mathrm{e}{-4}$ and batch size of $128$. We train it for 68 epochs with a learning rate schedule decay of $0.1$ at $15\textsuperscript{th}$, $35\textsuperscript{th}$ and $0.5$ at $50\textsuperscript{th}$ epoch.
\newline
\textit{PGD-7\cite{madrypgd}:} We use the training configuration as mentioned in \cite{model_cifar} to perform $7$-step adversarial training with $\epsilon=8/255$ and step size $2.5/255$.
\newline
\textit{IG Norm and IG-Sum Norm \cite{robust_attr_nips_sal} : } We report the accuracy as mentioned in the paper \cite{robust_attr_nips_sal}.
\newline
\textit{ART:} We use the same training configuration as mentioned for \textit{Natural} model with $\lambda = 0.5$ and $\beta = 50$. We calculate $\Tilde{x}$ using $\epsilon=8/255$, step size $1.5/255 $ and number of steps $a = 3$.  

\subsubsection{Attack Methodology and Evaluation}\label{section::attackmethodology}\hspace*{\fill} \\
For evaluation, we perform the Top-K variant of Iterative Feautre Importance Attack (IFIA) proposed by \cite{aaai_sal}. Feature importance function is taken as Integrated Gradients \cite{attr2017integrated}, and dissimilarity function is Kendall Correlation. The hyperparameters used are the same as in \cite{robust_attr_nips_sal} i.e. for CIFAR-10, SVHN and GTSRB datasets, $k$ in top-k is $100$, $\epsilon$ is $8/255$, number of steps is $50$ and step-size is $1/255$. For the Flowers dataset, $k$ is $1000$, $\epsilon$ is $8/255$, number of steps is $100$ and step-size is $1/255$. We also show the comparison by varying $\epsilon$ on CIFAR-10 dataset in Section \ref{sec::additional_attr}. Evaluation is also similar to \cite{robust_attr_nips_sal} using Top-k intersection and Kendall correlation measure and we report both numbers as percentage values. For Top-k intersection, $k$ is $100$ for CIFAR-10, SVHN and GTSRB datasets, and $1000$ for Flowers dataset.  

\subsection{Additional Analysis on CIFAR-10}\label{sec::additional_attr}

\begin{figure}[t]
\centering
    \begin{minipage}{.45\textwidth}
      \includegraphics[width=\linewidth,height=0.5\linewidth]{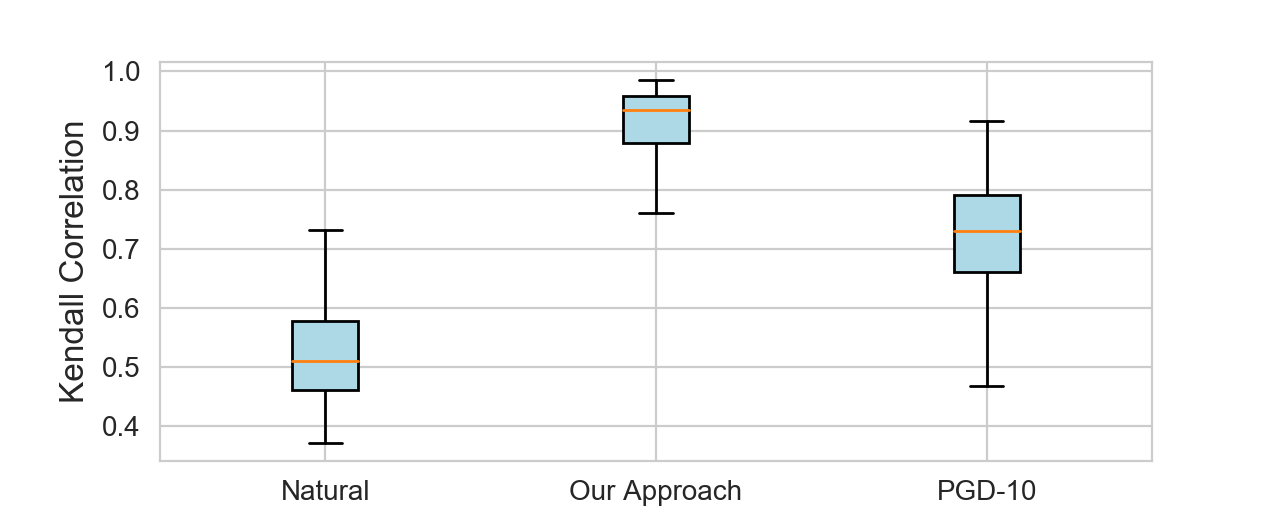}
    \end{minipage}
    \begin{minipage}{.45\textwidth}
      \includegraphics[width=\linewidth,height=0.5\linewidth]{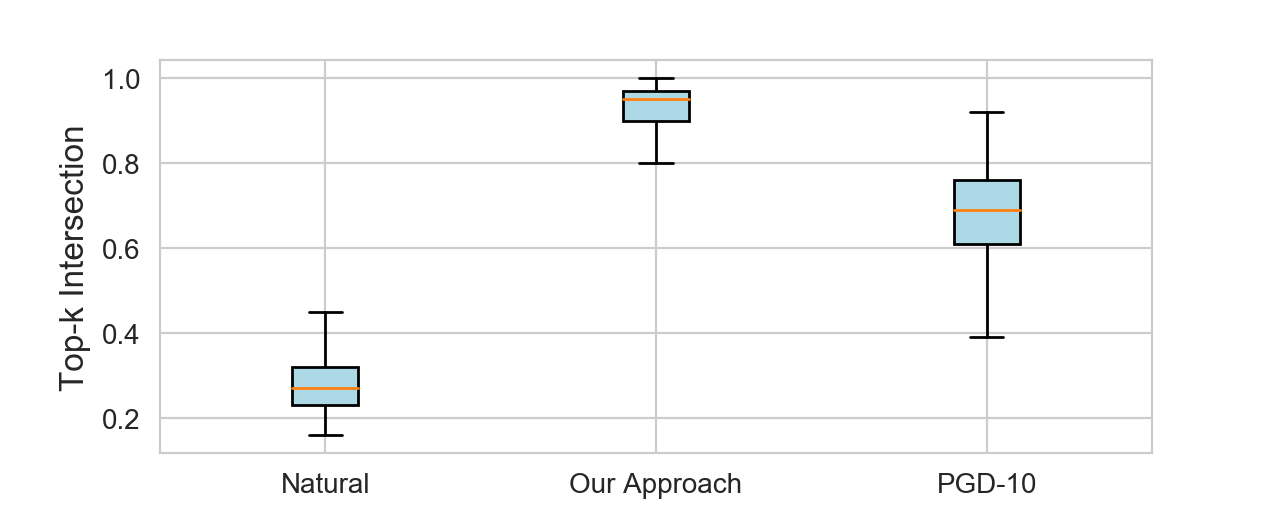}
    \end{minipage}
\caption{\footnotesize{Variance box plot of Attributional Robustness measure for different models on Kendall Correlation (left) and Top-k Intersection (right) for $1000$ test samples of CIFAR-10}}
\label{fig:var_plot_eval}
\end{figure}

\begin{figure}[t]
\centering
\scalebox{0.8}{
    \includegraphics[width=0.9\linewidth , height=0.4\linewidth]{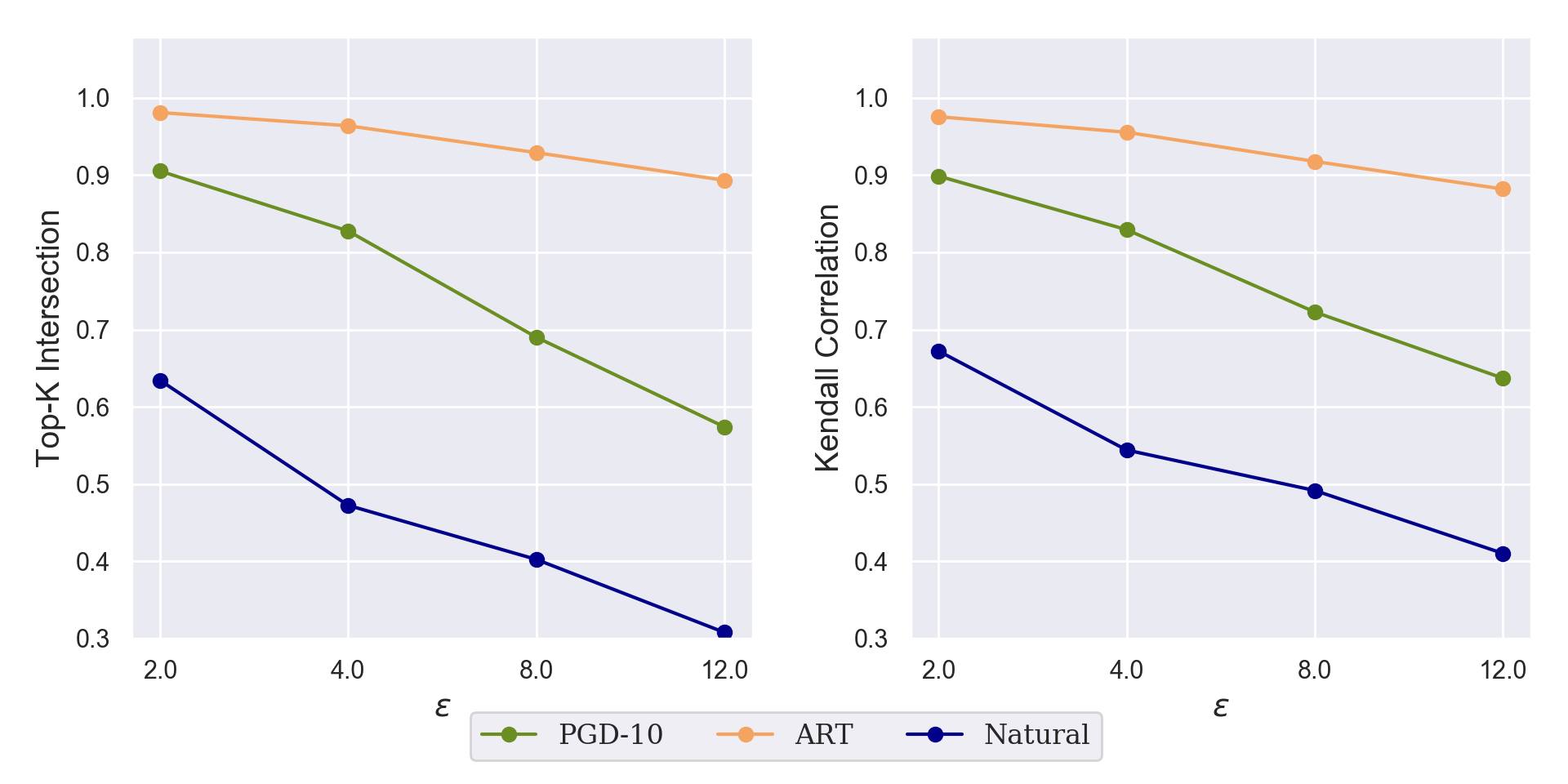}
    }
\caption{\footnotesize{Attributional robustness on varying $\epsilon$ for \textit{ART}, \textit{PGD-10} and \textit{Natural} models on CIFAR-10}}
\label{fig:art_acc_eps}
\end{figure}






\paragraph{Attributional Robustness:}\label{additional_attr}
In Fig \ref{fig:var_plot_eval}, we show the variance box plot of Kendall Correlation and Top-k Intersection with $\epsilon=8/255$ for \textit{Natural}, \textit{ART} and \textit{PGD-10} \cite{madrypgd} models on CIFAR-10. \textit{ART} has higher attributional robustness with the least variance as compared to other approaches across $1000$ samples randomly selected from the test dataset. We also measure the attributional robustness of models on varying $\epsilon$ to the standard values of $2/255$, $4/255$, $8/255$ and $12/255$ in the attack methodology as explained in Section \ref{section::attackmethodology}. Figure \ref{fig:art_acc_eps} shows the Top-k Intersection and Kendall correlation measure for the same. We can see that \textit{ART} outperforms \textit{PGD-10} and \textit{Natural} model over all choices of $\epsilon$. 



\begin{figure*}[t]
\centering
    \begin{minipage}{.24\textwidth}
      \includegraphics[width=\linewidth]{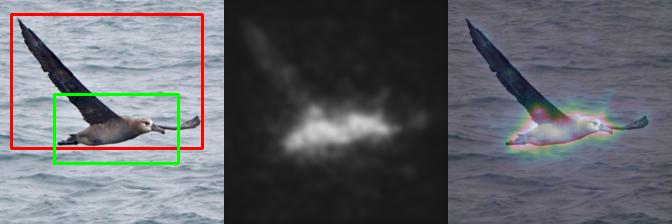}
    \end{minipage}
    \begin{minipage}{.24\textwidth}
      \includegraphics[width=\linewidth]{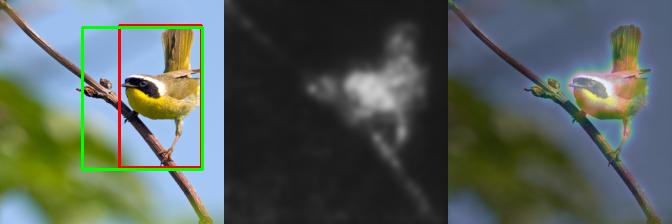}
    \end{minipage}
    \begin{minipage}{.24\textwidth}
      \includegraphics[width=\linewidth]{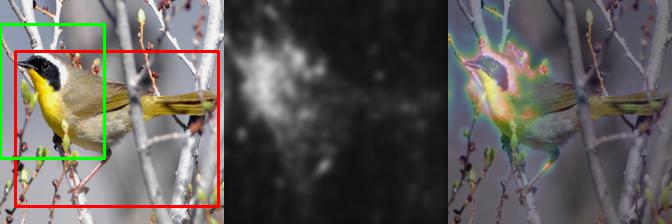}
    \end{minipage}
    \begin{minipage}{.24\textwidth}
      \includegraphics[width=\linewidth]{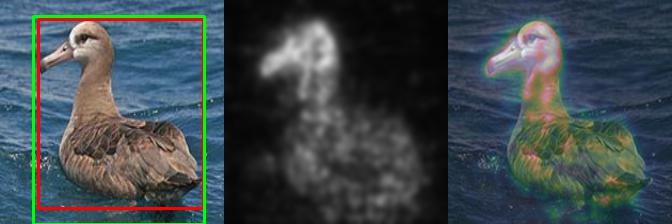}
    \end{minipage}\\
    \begin{minipage}{.24\textwidth}
      \includegraphics[width=\linewidth]{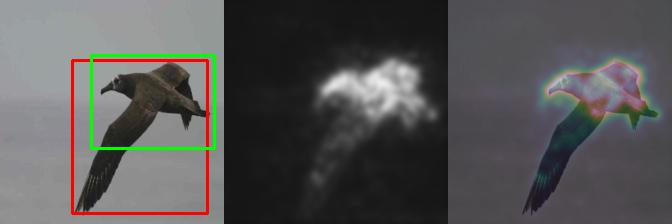}
    \end{minipage}
    \begin{minipage}{.24\textwidth}
      \includegraphics[width=\linewidth]{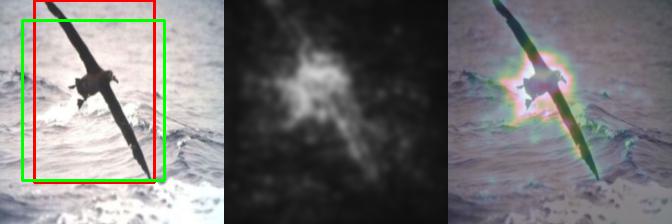}
    \end{minipage}
    \begin{minipage}{.24\textwidth}
      \includegraphics[width=\linewidth]{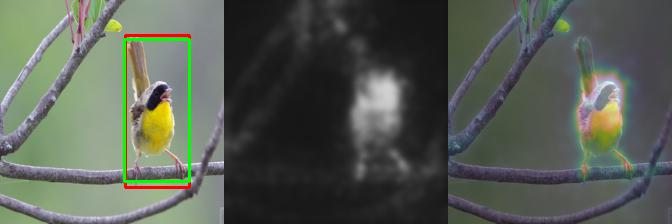}
    \end{minipage}
    \begin{minipage}{.24\textwidth}
      \includegraphics[width=\linewidth]{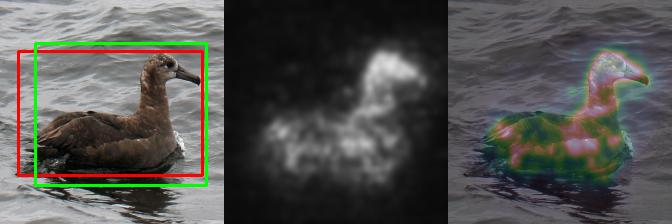}
    \end{minipage}\\
    \begin{minipage}{.24\textwidth}
      \includegraphics[width=\linewidth]{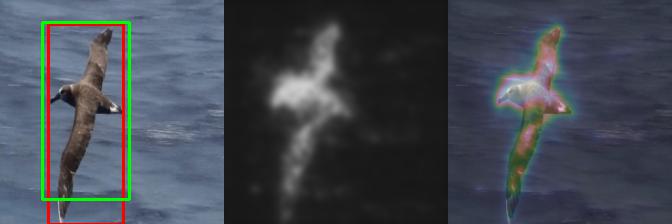}
    \end{minipage}
    \begin{minipage}{.24\textwidth}
      \includegraphics[width=\linewidth]{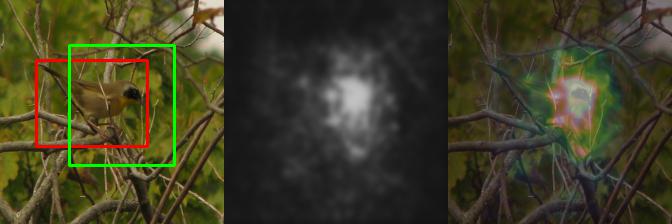}
    \end{minipage}
    \begin{minipage}{.24\textwidth}
      \includegraphics[width=\linewidth]{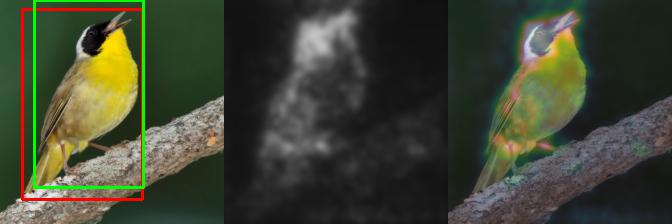}
    \end{minipage}
    \begin{minipage}{.24\textwidth}
      \includegraphics[width=\linewidth]{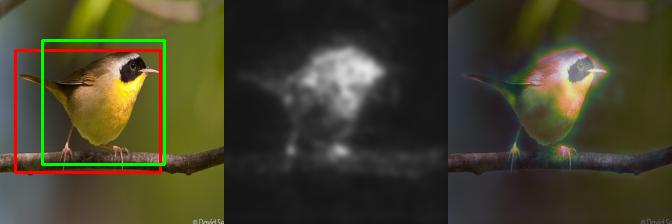}
    \end{minipage}
    \caption{\footnotesize{Examples of estimated bounding box and heatmap by ResNet50 model trained via our approach on randomly chosen images of CUB dataset; Red bounding box is ground truth and green bounding box corresponds to the estimated box}}
    \label{fig:loc_our}
\end{figure*}

\section{Weakly Supervised Localization: More Details and Results}\label{wsol_additional}
\label{sec_wsol_addl_results}
In this section, we provide more details of the dataset used for the results presented in the main paper on weakly supervised localization (Section 4.2), as well as more qualitative examples for these experiments.

\subsection{Dataset and Implementation Details}\label{wsol_dataset}
We begin by describing the dataset used in experiments for weakly supervised localization, which we could not include in the main paper owing to space constraints.

\noindent \underline{Dataset and Model:} CUB-200 \cite{cub} is an image dataset of 200 different bird species (mostly North American) with $11,788$ images in total. The information as a bounding box around each bird is also available. We finetune a ResNet-50 \cite{resnet} model pre-trained on ImageNet for the reported approaches as in \cite{adl}.

\noindent \underline{Hyper-parameters for training} \newline
\textit{Natural:} We use SGD optimizer with an initial learning rate of $0.01$, momentum of $0.9$ and $l_2$ weight decay of $1\mathrm{e}{-4}$. We train the model for $200$ epochs with batch size $128$ and learning rate decay of $0.1$ at every $60$ epochs.\newline
\textit{PGD-7 \cite{madrypgd}:} We use same hyper-parameters as natural training with $\epsilon=2/255.$ and $ step\_size=0.5/255.$ for calculating adversarial examples. \newline
\textit{ART:} We use SGD optimizer with an initial learning rate of 0.01, momentum of $0.9$ and $l_2$ weight decay of $1\mathrm{e}{-4}$. We decay the learning rate by $0.1$ at every $40$ epoch till $200$ epochs and train with a batch size of $90$. While calculating $L_{attr}$ loss, we took mean over channels of images and gradients. Values of other hyper-parameters are $\epsilon=2/255$, $step\_size=1.5/255$, $a =3$, $\lambda = 0.5$ and $\beta = 50$. 

\subsection{Qualitative Analysis}
Figure \ref{fig:loc_our} shows the estimated bounding box and heatmap derived from gradient based attribution \cite{attr2013gradient} on randomly sampled images for ResNet50 model trained via our approach. We observe that the estimated bounding box sometimes does not capture the complete object in cases where birds have extended wings, or the bird is in an occluded area with branches and twigs. Although, we observe qualitatively that this issue also exists for other models \cite{adl}.

\begin{figure}[t]
\centering
\scalebox{0.8}{
    \includegraphics[width=0.9\linewidth , height=0.4\linewidth]{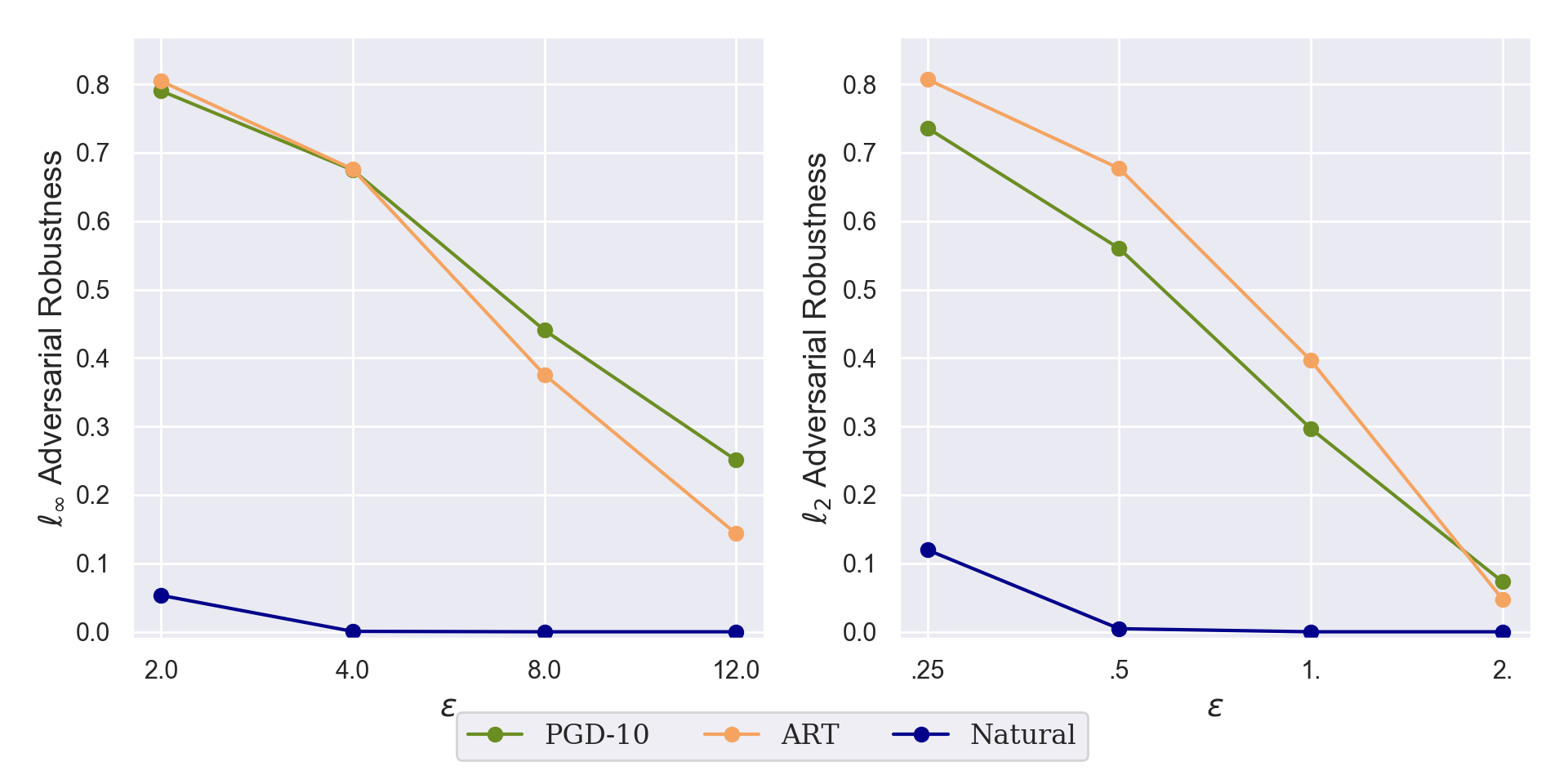}}
\caption{\footnotesize{$\ell_{\infty}$ and $\ell_{2}$ adversarial robustness on varying $\epsilon$ of  \textit{ART}, \textit{PGD-10} and  \textit{Natural} model on CIFAR-10}}
\label{fig:adv_acc_eps}
\end{figure}

\begin{table}[t]
\centering
\caption{\footnotesize{Comparison of Adversarial accuracy of different baseline models using transfer-based black-box attacks on CIFAR-10 }}
\begin{tabular}{|c|c|c|c|c|}
\hline
\multirow{2}{*}{Training Approach} & \multicolumn{3}{c|}{\begin{tabular}[c]{@{}c@{}}Adversarial perturbation\\ created using\end{tabular}} & \multirow{2}{*}{\begin{tabular}[c]{@{}c@{}}Clean Test \\ Accuracy\end{tabular}} \\
                                   & Natural                           & PGD-10                           & ART                            &                                                                                 \\ \hline
Natural                            & 0.00                              & 80.35                            & 49.09                          & 95.26                                                                           \\
PGD-10                             & 86.44                             & 44.07                            & 71.34                          & 87.32                                                                           \\
ART                                & 88.45                             & 72.72                            & 37.58                          & 89.84                                                                           \\ \hline
\end{tabular}
\label{transfer_adv}

\end{table}

\section{Adversarial Robustness}
\label{sec_adv_robustness_ablations}
In this section, we provide additional results on adversarial robustness on the CIFAR-10 dataset.

\paragraph{\textbf{Adversarial Robustness on $\ell_\infty$ and $\ell_2$ PGD Perturbations with Varying $\epsilon$}}
To analyze the adversarial robustness of \textit{ART} model, we report and compare accuracy of the \textit{ART} model and the \textit{PGD-10} adversarially trained model over $\ell_\infty$ and $\ell_2$ PGD perturbations for different values of $\epsilon$ on CIFAR-10. In Figure \ref{fig:adv_acc_eps}, we can observe that \textit{ART} adversarial robustness for $\ell_\infty$ perturbations is similar to \textit{PGD-10} for $\epsilon$ less than $4/255$ and better for various values of $\ell_{2}$ perturbations.

\paragraph{\textbf{Transfer-based black-box attacks}}
We analyse the adversarial robustness of \textit{ART} models on transfer-based black box attacks. Specifically, we compute the adversarial perturbations on the test set of CIFAR-10 for different baseline models and evaluate its adversarial accuracy on \textit{ART}. We see that the transfer of adversarial perturbation from \textit{ART} is much better than PGD-10 on Natural model. \textit{ART} also shows higher robustness than PGD-10 for transfer attack from Natural model as reported in table \ref{transfer_adv}.

\paragraph{\textbf{Comparison with other training techniques for adversarial robustness:}}
We consider \textit{JARN}\cite{jarn_iclr} and \textit{CURE}\cite{cure}, which are recently proposed training techniques for adversarial robustness that are different from adversarial training \cite{madrypgd}. We compare the adversarial robustness of these techniques with \textit{ART} on CIFAR-10 dataset using a ${\ell}_{\infty}$ PGD-20 adversarial perturbation with $\epsilon=8/255$. \textit{JARN}, \textit{CURE} and \textit{ART} show adversarial accuracy of $15.5\%$, $41.4\%$ and $37.73\%$ respectively and test accuracy of $93.9\%$, $83.1\%$ and $89.84\%$ respectively. 

\paragraph{\textbf{Using $L_{attr} + L_{ce}$ to Compute Perturbations $\Tilde{x}$}}
With the motive to combine the benefits from attributional and adversarial robust models, we augment the loss function of our approach with adversarial loss \cite{madrypgd}.
We observe that the model achieves test accuracy of $85.33$ and adversarial accuracy of $52.31$ on PGD-40 $\ell_{\infty}$ attack with $\epsilon=8/255$ as compared to the \textit{PGD-10} model which has $87.32$ test accuracy and $44.07$ adversarial accuracy. The attributional robustness measure of Top-k intersection and kendall correlation using Integrated Gradients is $74.24$ and $77.86$ which is less than the attributional robustness of \textit{ART} model but is $\sim 5 \%$ better than \textit{PGD-10} model. 
\end{document}